\documentclass[dvipsnames]{article}

\usepackage[square,sort,comma,numbers]{natbib}

\usepackage[final]{neurips_data_2022}

\usepackage{hyperref}       
\usepackage{url}            
\usepackage{booktabs}       
\usepackage[table]{xcolor}         
\usepackage{graphicx}
\usepackage{algorithm,algpseudocode}
\usepackage{amsmath}
\usepackage[inline,shortlabels]{enumitem}
\usepackage{subcaption}
\usepackage{diagbox}
\newcolumntype{P}[1]{>{\centering\arraybackslash}p{#1}}
\graphicspath{{figures/}}

\makeatletter
\def\Url@twoslashes{\mathchar`\/\@ifnextchar/{\kern-.2em}{}}
\g@addto@macro\UrlSpecials{\do\/{\Url@twoslashes}}
\makeatother

\newcommand{\revklim}[1]{#1}

\newcommand{\hideme}[1]{}

\algnewcommand{\LeftComment}[1]{\Statex \(\triangleright\) #1}
\newcommand{\ouraida}{AIDA$^{+}$}

\newcommand{\eg}{e.g., }

\newcommand{\ie}{i.e., }

\newcommand{\figref}[1]{Fig.~\ref{#1}}    
\newcommand{\figsref}[2]{Figs.~\ref{#1}--\ref{#2}}    
\newcommand{\Figref}[1]{Figure~\ref{#1}}  
\newcommand{\tabref}[1]{Table~\ref{#1}}
\newcommand{\Tabref}[1]{Table~\ref{#1}}

\newcommand{\secref}[1]{Section~\ref{#1}}
\newcommand{\equref}[1]{Eq.~(\ref{#1})}

\newcommand{\ourdataset}{TempEL}

\definecolor{deepcarmine}{rgb}{0.93, 0.47, 0.13}
\colorlet{negative0001}{deepcarmine!75}
\colorlet{negative001}{deepcarmine!50}
\colorlet{negative01}{deepcarmine!25}

\definecolor{darkspringgreen}{rgb}{0.0, 0.5, 1.0}
\colorlet{positive0001}{darkspringgreen!75}
\colorlet{positive001}{darkspringgreen!50}
\colorlet{positive01}{darkspringgreen!25}

\newcommand{\authGhent}{$\natural$}
\newcommand{\authCPH}{$\sharp$}

\title{{\ourdataset}: Linking Dynamically Evolving and\\ Newly Emerging Entities}

\author{Klim Zaporojets\textsuperscript{\,\authGhent}\hspace{0.5cm} Lucie-Aimée Kaffee\textsuperscript{\,\authCPH}\hspace{0.5cm} Johannes Deleu\textsuperscript{\,\authGhent}\\
\textbf{Thomas Demeester}\textsuperscript{\,\authGhent}\hspace{0.5cm} \textbf{Chris Develder}\textsuperscript{\,\authGhent}\hspace{0.5cm} \textbf{Isabelle Augenstein}\textsuperscript{\,\authCPH}\\
  \textsuperscript{\authGhent}\,Ghent University -- imec, IDLab, Ghent, Belgium \\
  \textsuperscript{\authCPH}\,Dept.\ of Computer Science, University of Copenhagen, Denmark \\
  \texttt{\{klim.zaporojets,johannes.deleu,thomas.demeester,chris.develder\}@ugent.be}\\
  \texttt{\{kaffee,augenstein\}@di.ku.dk}\\
  }

\begin{document}

\maketitle

\begin{abstract}

In our continuously evolving world, entities change over time and new, previously non-existing or unknown, entities appear.
    We study how this evolutionary scenario impacts the performance 
    on
    a well established \textit{entity linking} (EL) task. 
    For that study, we
    introduce {\ourdataset}, an entity linking dataset that consists of time-stratified English Wikipedia snapshots from 2013 to 2022, 
    from which we collect both \emph{anchor mentions} of entities, and these \emph{target entities}' descriptions.
    By capturing such temporal aspects, 
    our newly introduced {\ourdataset} resource
    contrasts with 
    currently 
    existing
    entity linking 
    datasets, which are 
    composed 
    of
    fixed mentions linked to a
    single static version of a target Knowledge Base (\eg Wikipedia 2010 for CoNLL-AIDA). 
    Indeed, for each of 
    our collected temporal snapshots, {\ourdataset} contains 
    links 
    to 
    entities that are \emph{continual}, \ie occur in 
    all of 
    the years,
    as well as 
    completely \emph{new} 
    entities that appear for the first time at some point.
    Thus, we enable to quantify the performance of 
    current state-of-the-art EL models for:
    \begin{enumerate*}[(i)]
        \item entities that are subject to changes over time in  their Knowledge Base 
        descriptions
        as well as their mentions' contexts, and
        \item newly created entities that were previously non-existing (\eg at the time the EL model was trained).
    \end{enumerate*}
    Our experimental results show that in terms of temporal performance degradation, 
    \begin{enumerate*}[(i)]
        \item \emph{continual} entities suffer a decrease of up to \revklim{3.1\% }
        EL accuracy, 
        while
        \item for \emph{new} entities this accuracy drop is up to \revklim{17.9\%}.
    \end{enumerate*}
    This highlights the challenge of the introduced {\ourdataset} dataset and opens new research prospects in the area of time-evolving entity disambiguation.\footnote{\revklim{\ourdataset~dataset, code and models are made public at \url{https://github.com/klimzaporojets/TempEL}.}}  
    
\end{abstract}

\section{Introduction}
\begin{figure}[!t]
\centering
\includegraphics[width=0.7\columnwidth]{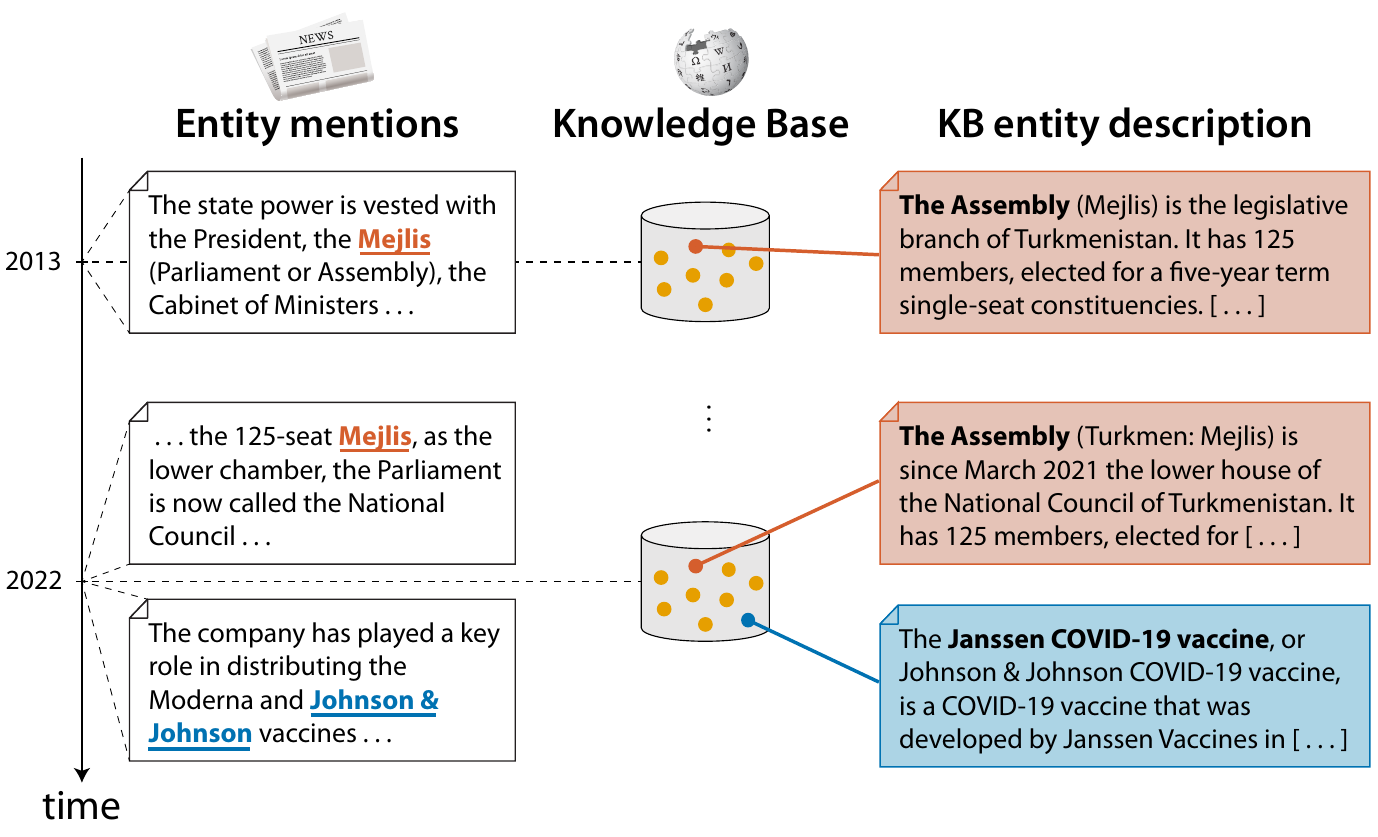}
\caption{Illustration of KB entities changing over time: the ``Mejlis'' entity changes over time (both in its KB description and the contexts in which it is referenced to), while the Johnson \& Johnson vaccine is an entirely new one that did not exist before.}
\label{fig:el_introduction}
\end{figure}

Entity linking (EL) is a well-established task that is concerned with mapping anchor \textit{mentions} in text to target \textit{entities} that 
describe
them in a Knowledge Base (KB) (\eg Wikipedia).\footnote{Some of the related work 
\citep{ganea2017deep, kolitsas2018end, sevgili2020neural, zhang2021entqa, zaporojets2021consistent} 
distinguishes between \textit{entity disambiguation} and \textit{entity linking} tasks. This latter including \textit{mention detection} and \textit{disambiguation} in an end-to-end setting. In the current work, we follow a more conservative 
naming convention 
\cite{rao2013entity, wu2019zero, logeswaran2019zero, onoe2020fine, raiman2022deeptype}, and use the term \textit{entity linking} and \textit{entity disambiguation} interchangeably.}
Existing benchmark datasets for EL \citep{usbeck2015gerbil,roder2018gerbil,sevgili2020neural,petroni2020kilt} are composed of a fixed set of annotated mentions linked to a single version of a target KB.
This static setup is oblivious to the inherently non-stationary nature of the entity linking task where both target entities as well as anchor mentions change over time. The example in \figref{fig:el_introduction} illustrates this time-evolving essence of entity linking with a simple evolutionary comparison between Wikipedia 2013 and 2022. It showcases two scenarios studied in the current paper: \begin{enumerate*}[(i)]
    \item temporal evolution of existing (\textit{continual}) entities across temporal snapshots, and
    \item appearance of \emph{new}, previously non-existent entities
\end{enumerate*}. 
For instance,
the description of the \emph{continual} entity \textit{The Assembly} differs between Wikipedia 2013 and 2022. Furthermore, the context of a mention ``Mejlis'' referring to \textit{The Assembly} also changes over time. Conversely, the \emph{new} entity \textit{Janssen COVID-19 vaccine} is newly introduced in 2021 with the corresponding mentions (\eg ``Johnson \& Johnson'' in \figref{fig:el_introduction}) that are linked to it. 

In this paper 
we introduce \ourdataset, a novel dataset to study this time-evolving aspect of the entity linking task.
We therefore extract 10 equally spread 
yearly snapshots  
from English Wikipedia 
entities starting from 
January 1, 2013 until January 1, 2022.
We use each of these temporal snapshots of Wikipedia to also extract anchor mentions with the surrounding text. Thus, {\ourdataset} captures the temporal evolution not only in the target entities as they are defined in 
the Wikipedia KB,
but also in the contexts of anchor mentions linked to these entities. 
Each of the 10 temporal snapshots of our dataset 
is composed of training, test and validation sets with equal numbers of mentions and entities 
across the 
snapshots. 
Furthermore, 
\ourdataset~is designed to comprise 
mentions pointing to \emph{continual} entities across all 
the temporal snapshots, and to \emph{new} entities inside a given temporal 
snapshot. 

Finally, as a baseline, we finetune and evaluate the bi-encoder component of the BLINK model \citep{wu2019zero} on the various temporal snapshots of our newly introduced {\ourdataset} dataset. 
The bi-encoder is widely used in state-of-the-art entity linking models \citep{zhang2021entqa, wu2019zero} to retrieve the top $K$ (in this work we experiment with $K = 64$)
candidate target entities for a given anchor mention context. Furthermore, its straightforward finetuning and fast retrieval performance on millions of candidate entities \citep{johnson2019billion}, make it an ideal choice to test on \ourdataset.
Our experiments demonstrate
a 
consistent
temporal model deterioration 
for mentions linked to both
\emph{continual} 
(\revklim{3.1\% }
accuracy@64
points)
as well as \emph{new} 
(\revklim{17.9\%} accuracy@64 points)
entities. A more detailed analysis
reveals
that the maximum drop in performance is observed for \emph{new} entities that require 
fundamentally different world knowledge that was not present in the corpus originally used to pre-train BERT. 
This is e.g. the case for \emph{new} entities related to COVID-19 for which the bi-encoder model suffers additional deterioration of 
\revklim{14\%}
accuracy@64
points compared to the rest of the new entities. 

\section{Related work}
\label{sec:related}

Our work is related to multiple different, yet interconnected research areas described below. First, we explain how \ourdataset~compares to the currently widely used \textit{entity linking datasets}. Next, we relate our work to already existing \textit{temporal datasets} covering different aspects of the temporal evolution of the data. Finally, we describe the existing \textit{entity-centric} research efforts, comparing the \ourdataset~entity linking dataset to other datasets that heavily depend on the use of entities.
\paragraph{Entity linking datasets} Most current state-of-the-art EL models \citep{yamada2020global, orr2020bootleg, de2020autoregressive, zhang2021entqa, de2021highly} report on 
datasets from predominantly the news domain such as AIDA \citep{hoffart2011robust}, 
KORE50~\citep{hoffart2011robust}, AQUAINT \citep{milne2008learning}, ACE 2004, MSNBC \citep{ratinov2011local}, N$^3$ \citep{roder2014n3}, DWIE\citep{zaporojets2021dwie}, VoxEL\citep{rosales2018voxel}, and TAC-KBP 2010-2015 \citep{ji2010overview,ji2015overview}. Other frequently used datasets include the web-based IITB \citep{kulkarni2009collective} and OKE 15/16 \citep{nuzzolese2015open}, as well as the tweet-based Derczynski \cite{derczynski2015analysis}. Additionally, larger yet automatically annotated datasets such as WNED-WIKI and WNED-CWEB \citep{guo2018robust} 
have been also widely adopted. 
Finally, 
a number of
resources such as the domain-specific biomedical MedMentions \citep{mohan2018medmentions}, the zero-shot ZeShEL \citep{logeswaran2019zero}, and the multi tasking DWIE \citep{zaporojets2021dwie} and \ouraida \citep{zaporojets2021consistent} datasets have been recently introduced. 
Many of the mentioned datasets are further covered by entity linking evaluation frameworks such as GERBIL \citep{usbeck2015gerbil,roder2018gerbil} and KILT \citep{petroni2020kilt} that provide a common interface to evaluate the models. Yet, the mentioned resources are limited to static mention annotations linked to entities from a single version of a Knowledge Base. 
This contrasts with our newly introduced {\ourdataset} dataset, where the anchor mentions as well as the target entity descriptions are taken from different time periods. The datasets most closely related to our work are the recently introduced WikilinksNED \citep{eshel2017named, onoe2020fine} and ShadowLink \citep{provatorova2021robustness}. WikilinksNED contains only unseen mention-entity pairs in its test subset, thus encouraging the design of models invariant to overfitting and memorization biases. Furthermore, ShadowLink contains \textit{overshadowed entities}: entities referred to by ambiguous mentions whose most likely target entity is different, \eg the anchor mention ``Michael Jordan'' linked to the scientist instead of to the more widely referred to target entity describing the former basketball player. We incorporate the challenges presented in both of these datasets in \ourdataset~(see \secref{sec:dataset_construction} for further details).
\paragraph{Temporal datasets} Research on temporal drift in data has gained a lot of interest in  recent  years. The focus has mostly been on creating datasets to train language models on different temporal snapshots of corpora derived from scientific~\citep{lazaridou2021mind}, newswire~\citep{lazaridou2021mind,dhingra2022time}, Wikipedia~\citep{jang2022temporalwiki}, and Twitter~\citep{loureiro2022timelms} domains. 
More recently, temporal datasets have appeared to address 
tasks 
such as sentiment analysis \citep{lukes2018sentiment, ni2019justifying, agarwal2022temporal}, text classification \citep{huang2018examining,he2018time}, named entity recognition \citep{derczynski2016broad,rijhwani2020temporally}, question answering \citep{lazaridou2021mind}, and entity typing \citep{luu2021time}, among others. 
However, the creation of datasets tackling the temporal aspect of entity linking has largely been left unexplored. 
To the best of our knowledge, the dataset most closely related to {\ourdataset} is diaNED, introduced by \cite{agarwal2018dianed}. There, the authors annotate mentions 
that require additional temporal information from the context to be correctly disambiguated. Conversely, in {\ourdataset} both mentions and entities are extracted from evolving temporal snapshots.
 
\paragraph{Entity-driven datasets} 
Recent research has demonstrated the benefits of incorporating entity knowledge in various downstream tasks \citep{yang2017leveraging, peters2019knowledge,yamada2020luke,guu2020realm,verga2021adaptable,yasunaga2021qa,liu2022knowledge}.
This progress has been accompanied by the creation of entity-driven datasets for tasks such as language modeling \citep{petroni2019language,agarwal2021knowledge,kassner2021multilingual}, question answering \citep{yih2015semantic, joshi2017triviaqa, jiang2019freebaseqa,lewis2021paq,saxena2021question}, fact checking \citep{thorne2018fever, onoe2021creak, aly2021feverous} and information extraction  
\citep{yao2019docred,zaporojets2021dwie}, 
to name a few. 
Yet, recent findings \citep{runge2020exploring,fevry2020entities,lewis2020retrieval,verlinden2021injecting,heinzerling2021language,ri2021mluke}
suggest that entity \textit{representation} and \textit{identification} 
(\ie identifying the correct entity that match a given text) 
are among the main challenges 
that should be solved to further increase performance on such datasets.
We believe that {\ourdataset} can contribute to addressing these challenges by:
\begin{enumerate*}[(i)]
\item encouraging research on devising more robust methods to creating \textit{entity representations} that are invariant to temporal changes; and 
\item improving entity identification for non-trivial scenarios involving ambiguous and uncommon mentions (\eg linked to \textit{overshadowed entities} as defined above).
\end{enumerate*} 
\section{The {\ourdataset} dataset}
In this section we will provide details on how {\ourdataset} was constructed (\secref{sec:dataset_construction}), describing the main components of the creation pipeline as sketched in \figref{fig:el_pipeline}.
Furthermore, we discuss the aspects taken into account to guarantee the overall quality of our dataset (\secref{sec:quality}). Finally, we present statistics of {\ourdataset} (\secref{sec:dataset_statistics}), 
illustrating its dynamically evolving nature. 
\subsection{Dataset construction}
\label{sec:dataset_construction}

\begin{figure}[!t]
\centering
\includegraphics[width=1.0\columnwidth]{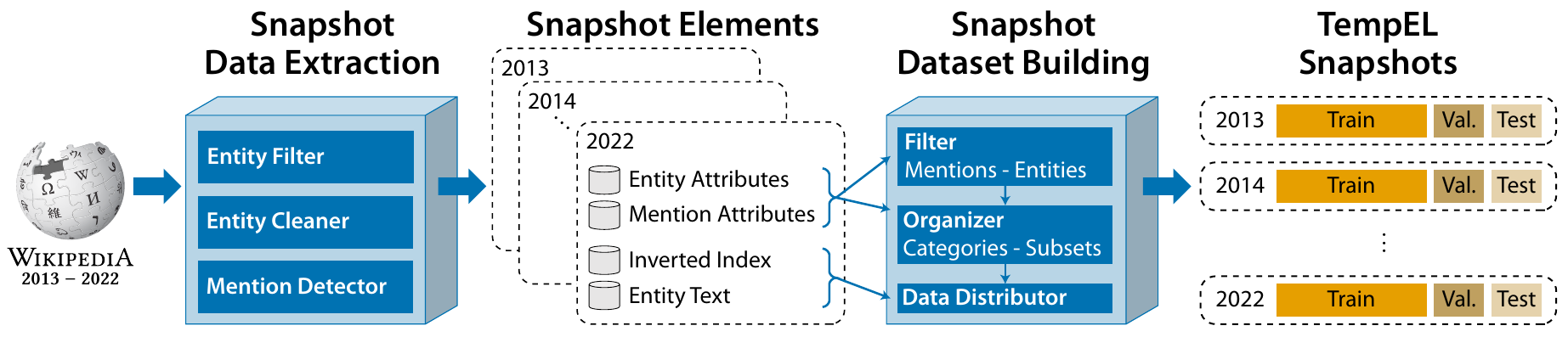}
\caption{The pipeline to create our {\ourdataset} dataset. All the components are explained in \secref{sec:dataset_construction}.}
\label{fig:el_pipeline}
\end{figure}

\paragraph{Snapshot Data Extraction}
As \figref{fig:el_pipeline} indicates, we start from the history log dumps from February 1, 2022 of Wikipedia itself.
We first filter these (see \textit{Entity Filter} in \figref{fig:el_pipeline})
to:
\begin{enumerate*}[(i)]
    \item 
    exclude
    pages that are irrelevant for {\ourdataset} (\ie categories, disambiguation pages, redirects and lists); and
    \item select 
    the most temporally stable 
    version of 
    a Wikipedia page from the last month of the year in order to avoid introducing 
    more volatile and 
    potentially corrupted content edits (see \secref{sec:quality} for further details).
\end{enumerate*}
Next, 
the Wikipedia pages are cleaned (see \emph{Entity Cleaner} in \figref{fig:el_pipeline}) by stripping from the Wikitext markup content.\footnote{\url{https://en.wikipedia.org/wiki/Help:Wikitext}} We use both regular expressions as well as the MediaWiki API for more difficult cases, such as the 
parsing
of some of the Wikitext templates.
Finally, we detect the mentions (see \textit{Mention Detector} in \figref{fig:el_pipeline}) in each of the Wikipedia entity pages, filtering out the ones that point to anchors (\ie subsections in Wikipedia pages), pages in languages other than English,
files, red links (\ie links pointing to not yet existing Wikipedia pages) and redirects. 

The output of the \textit{Snapshot Data Extraction} step first of all includes a set of \emph{Entity} and \emph{Mention Attributes} (\eg the last modification date of the target entity), which are detailed in the supplementary material \revklim{(see \secref{app:mention_entity_attributes})}.
These attributes form part of the final dataset, making it possible to perform additional analyses of the results. 
Furthermore, the \emph{Inverted Index} is generated to quickly access the Wikipedia pages that include a mention for a given target entity.
Finally, the \emph{Entity Text} files are extracted containing the (potentially yearly varying) textual content from the Wikipedia entity definition, as well as anchor mentions therein. These mentions of Wikipedia anchors that link to an entity will be extracted in the \textit{Snapshot Dataset Building} step described further.

\paragraph{Snapshot Dataset Building} 
Starting from the \emph{Snapshot Elements} produced by the \emph{Snapshot Data Extraction} process described above, the actual 
\ourdataset~dataset is now generated.
The first step is to apply an additional \textit{Filter} to both entities and mentions with the goal of creating a more challenging dataset. 
This is done by
excluding mentions
for which the correct entity it refers to has the highest prior~\citep{yamada2016joint}.
More formally, the \textit{mention prior} is calculated as follows,  
    \begin{equation}
    P(e\vert m) = \vert A_{e,m} \vert / \vert A_{*,m} \vert \label{eq:mention_prior},   
    \end{equation}
where $A_{*,m}$ is the set of all anchors that have the same mention $m$, and $A_{e,m}$ is the subset thereof that links to entity $e$.
Additionally, 
we exclude the mentions whose normalized edit distance from the target entity title is below an established threshold.\footnote{During the generation of \ourdataset, we use a 
threshold of 0.2.} 
By ignoring the mentions with the highest prior and exact match with the title, we ensure that {\ourdataset} contains non-trivial disambiguation cases where the naive approaches (\eg defaulting to the most frequently linked entity for a given mention) would fail \citep{guo2018robust, logeswaran2019zero, wu2019zero, provatorova2021robustness}. 

Furthermore, the entities are organized (see \emph{Organizer} in \figref{fig:el_pipeline}) into two \emph{categories}: \begin{enumerate*}[(i)]
    \item \emph{new}, emerging and previously non-existent entities that are introduced in a particular snapshot; and
    \item \emph{continual} entities across all the temporal snapshots. 
\end{enumerate*} 
Next, the mentions are divided in separate subsets (\ie train, validation and test), with the constraint of normalized edit distance between the mentions in different subsets referring to the same target entity be higher than 0.2.
This way, we expect to discourage potential models from memorizing the mapping between mentions and entities \citep{onoe2020fine}. 

Finally, the data is distributed equally (see \emph{Data Distributor} in \figref{fig:el_pipeline}) across all of the temporal snapshots.
This way, the difference in performance can only be attributed to temporal evolution and not to 
inconsistencies
related to dataset variability (\eg different number of training instances in each of the temporal snapshots). Concretely, we enforce that the number of \emph{continual} and \emph{new} entities as well as the number of mentions stays the same across the temporal snapshots (see \tabref{tab:dataset_detail2}). 
We achieve this by performing a random mention subsampling in 
snapshots
with higher number of mentions, weighted by the difference in the number of mentions-per-entity. 
This produces
a very similar
mention-entity distribution 
across the temporal snapshots.
Finally, the filtered anchor mentions are located in the cleaned Wikipedia pages (\ie the \textit{Entity Text} in \figref{fig:el_pipeline}) using the \textit{Inverted Index} created in the previous \emph{Snapshot Data Extraction} step. 
The context of each of the mentions is further paired with the respective content of target pages, outputting this way the final {\ourdataset} dataset.
\begin{table}[t]
\caption{Summary statistics of \ourdataset. The number of entities and mentions is the same across all of the temporal snapshots.}
\label{tab:dataset_detail2}
\vspace{.75\baselineskip}
\centering
\begin{tabular}{l ccc}
        \toprule
         Statistic & \multicolumn{1}{c}{Train} & \multicolumn{1}{c}{Validation} & \multicolumn{1}{c}{Test} \\ 
         \midrule 
         Temporal Snapshots & 10 & 10 & 10 \\ 
         Continual Entities & 10,000 & 10,000 & 10,000 \\
         \ \ \ \# Anchor Mentions & 136,227 & 42,096 & 46,765 \\ 
        New Entities & 373 & 373 & 373 \\
         \ \ \ \# Anchor Mentions & 1,764 & 1,231 & 1,450 \\
        \bottomrule 
        \end{tabular}
\end{table}
\subsection{Quality control} 
\label{sec:quality}
\paragraph{Corrupted content} Wikipedia is an open resource that relies on efforts of millions of Wikipedians 
to update and extend its contents.\footnote{\url{https://en.wikipedia.org/wiki/Wikipedia:Wikipedians}} 
As such, 
that content is not always reliable, with errors due to human mistakes or intentional vandalism. 
Despite 
efforts to prevent the introduction of such erroneous 
edits
\citep{west2010detecting,dang2016quality,wang2020assessing}, 
we have detected numerous cases
of corrupted 
entity descriptions
during our preliminary tests.
As a result, we adopted a simple, yet very effective heuristic: for each of the entities of a particular yearly snapshot, we select the most \textit{stable} (\ie the version of the entity that lasted the longest before being changed) content 
of the last month of the year (December). Due to the fact that most of the corrupted content is rolled back very quickly, and even automatically by specialized bots \citep{zheng2019roles,jiang2020good}, this heuristic is very robust. We double checked the correctness of the extracted content by manually inspecting the evolution of hundred entities with lowest Jaccard vocabulary similarity between temporal snapshots and observed no obviously erroneous entries.
\paragraph{Entity relevance} 
We filter out entities that have less than 10 in-links (\ie number of mentions linking to the entity) or contain less than 10 tokens in its Wikipedia page in order to avoid including noisy content \citep{eshel2017named}. Additionally, in order to avoid evaluation bias towards mentions pointing to more popular entities 
\citep{orr2020bootleg,chen2021evaluating}, we 
limit
the number of mentions per entity to 10 for our test and validation sets. 
This way, 
we expect the accuracy scores to not be
dominated by links to 
popular target entities (\ie entities with a big number of incoming links). 

\paragraph{Content filtering} 
We only consider mentions linked to the main Wikipedia articles describing entities. The mentions pointing to anchors (subsections in a Wikipedia document), images, files, and wiki pages in other languages are filtered out in \textit{Snapshot Data Extraction} step (see \figref{fig:el_pipeline}). In this step we also ignore pages that are not Wikipedia articles (\eg files, information on Wikipedia users, etc.) as well as redirect pages. This way, the target entities as well as anchor mentions in our dataset are obtained from a cleaned list of candidate pages referring to entities that contain a meaningful textual description in Wikipedia. 

\paragraph{Dataset distribution} During the construction of \ourdataset, we constrain the subsets to be of equal size and contain 
similar
mention-per-entity distributions across all the temporal 
snapshots.
This is implemented in \textit{Data Distributor} sub-component of the dataset creation pipeline (see \secref{sec:dataset_construction}). For example, the number of mentions linked to continual entities in our training subset is 
\revklim{136,227}
across all of the 
snapshots
(see \tabref{tab:dataset_detail2} for further details). 
We argue that this setting will produce  uniform, structurally unbiased 
snapshots.
This will allow to study exclusively the temporal effect on the performance of the models for each of the different time periods.  
Our reasoning is supported by previous work demonstrating that the size alone of the training set \citep{loureiro2022timelms} as well as a different distribution of the number of mentions per entity \citep{orr2020bootleg} can significantly affect the performance of the final model. Furthermore, we do not constrain the total number of entities from the Wikipedia KB to be equal 
across the temporal snapshots
(see \figref{fig:nr_wiki_entities}), since we consider it a part of the evolutionary nature of the entity linking task (\ie the temporal evolution of the target KB) we intend to study. 
\paragraph{Flexibility and extensibility} Finally, we provide a framework that can be used to re-generate the dataset with different parameters as well as to extend it with newer temporal snapshots. 
This includes the option to generate a new dataset with a 
customized
number of temporal snapshots (\eg quarterly instead of yearly spaced), different mention attributes (\eg filtering by mention prior values), entity popularity (\eg filtering out entities that have more than a certain number of in-links), among others 
(see \revklim{\secref{app:hyperparameters}} of the supplementary material for a complete list). 
\begin{figure}[t]
\begin{subfigure}{.31\textwidth}
  \centering
  \includegraphics[width=1.0\linewidth, trim={0.0cm 0.0cm 0.0cm 1.00cm},clip]{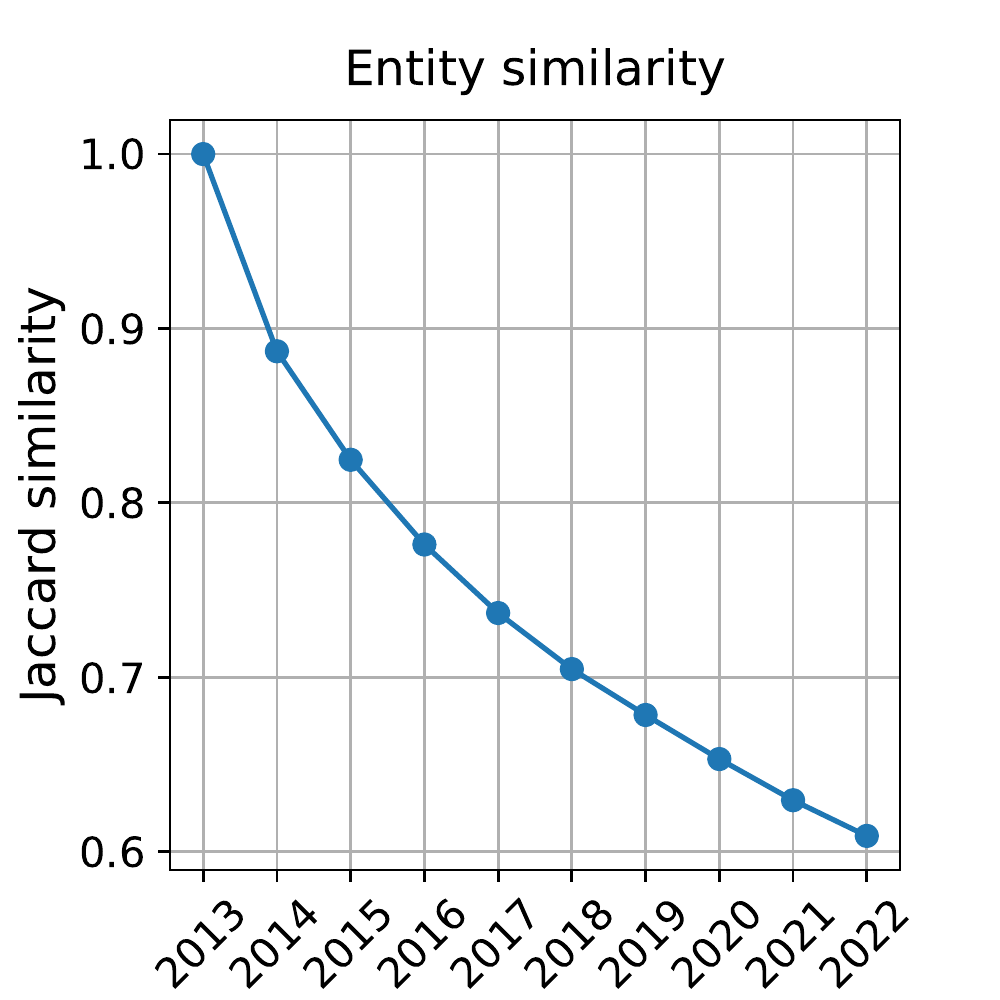}  
  \caption{Evolution of entities in terms of Jaccard vocabulary similarity.}
  \label{fig:evolution_entities_jaccard}
\end{subfigure} \hfill
\begin{subfigure}{.31\textwidth}
  \centering
  \includegraphics[width=1.0\linewidth, trim={0.0cm 0.0cm 0.0cm 1.00cm},clip]{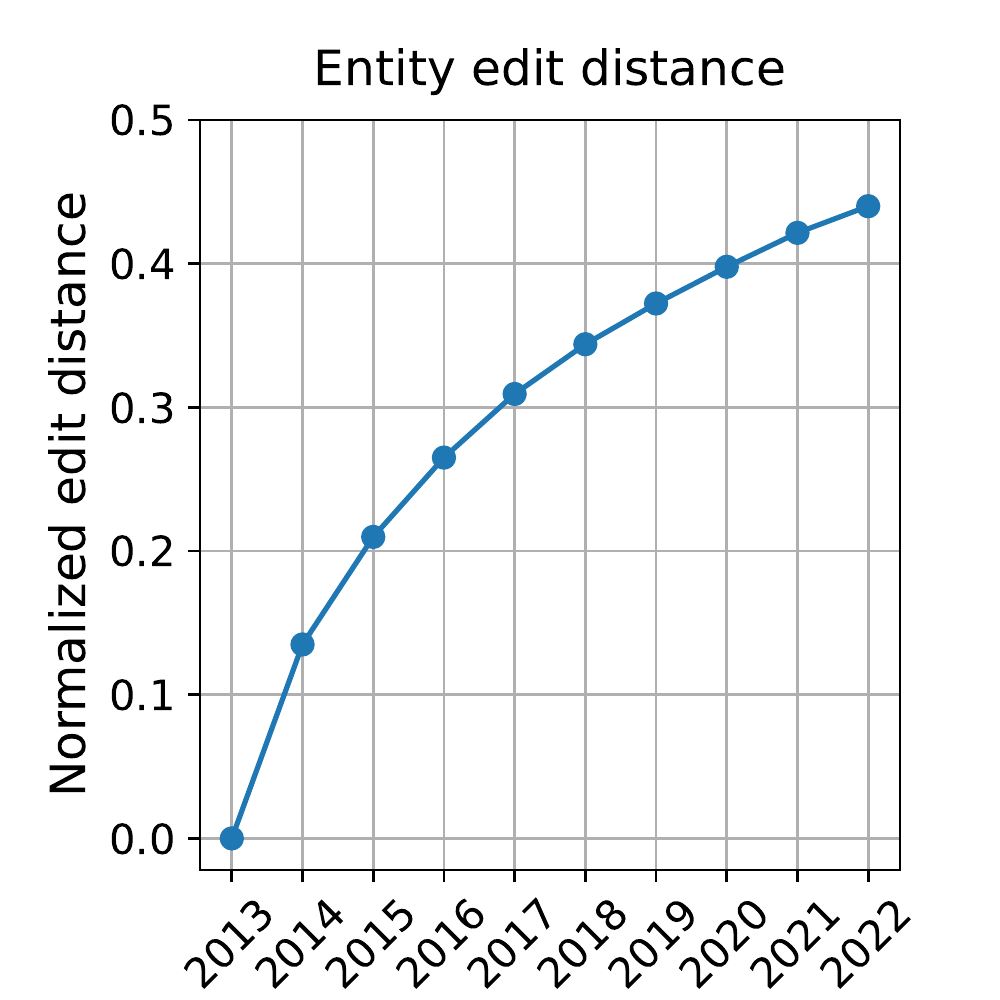}  
  \caption{Evolution of entities in terms of edit distance of the content.}
  \label{fig:evolution_entities_edistance}
\end{subfigure} \hfill
\begin{subfigure}{.31\textwidth}
  \centering
  \includegraphics[width=1.0\linewidth, trim={0.0cm 0.0cm 0.0cm 1.00cm},clip]{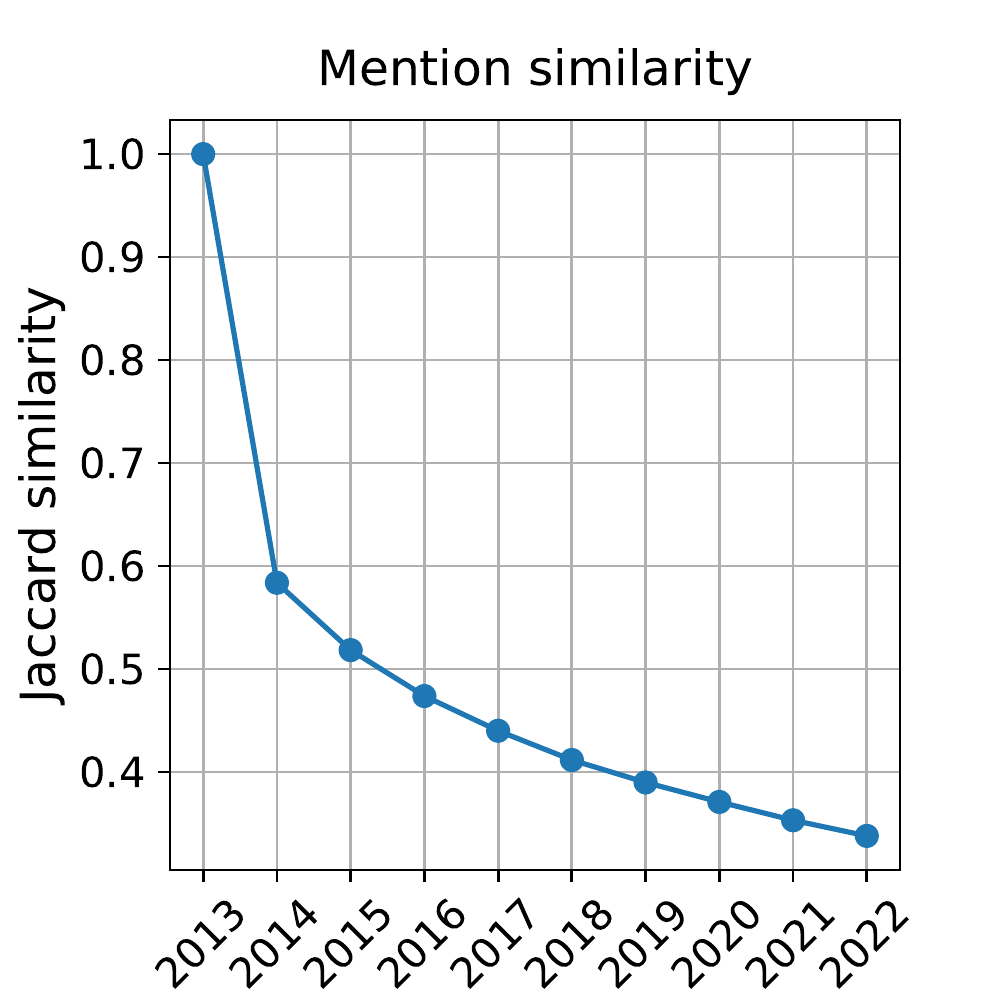}  
  \caption{Evolution of context around the mentions (Jaccard similarity).}
  \label{fig:evolution_mentions}
\end{subfigure}
\caption{Change of textual content of entities and context around mentions across temporal yearly snapshots (x-axis). }
\label{fig:fig}
\end{figure}
\subsection{Dataset statistics}
\label{sec:dataset_statistics}
\tabref{tab:dataset_detail2} summarizes the dataset statistics. 
We divide each of the temporal snapshots into train, 
validation and test subsets containing an equal number of \emph{continual} and \emph{new} entities. The number of mentions differs between the subsets since we
limit the number of mentions per entity to 10 in both validation and test sets
(see \textit{entity relevance} in \secref{sec:quality} for 
further
details). 

Additionally, we collect statistics related to temporal drift in content for both the target entities (Figs.~\ref{fig:evolution_entities_jaccard} and \ref{fig:evolution_entities_edistance}) as well as the context around the anchor mentions (\figref{fig:evolution_mentions}). Concretely, \figref{fig:evolution_entities_jaccard} visualizes Jaccard vocabulary similarity between the textual description of \emph{continual} entities in 2013 and that of posterior yearly snapshots
in \ourdataset. We observe a continual decrease, indicating that on average, the content of the entity description in Wikipedia is constantly evolving in terms of the used vocabulary. This is also supported by the graph in \figref{fig:evolution_entities_edistance}, which 
showcases a continuous
temporal increase of the average value of normalized edit distance across \emph{continual} entities. Finally, \figref{fig:evolution_mentions}  
illustrates
the temporal drift in the vocabulary (\ie Jaccard vocabulary similarity) of the context around the mentions pointing to 
the same entity.
We find it experiences a more significant change compared to the Jaccard similarity of entity content illustrated in \figref{fig:evolution_entities_jaccard}. This suggests that the context around the anchor mentions 
is subject to
a higher degree of temporal 
transformation
compared to that of target entities, making it an interesting 
item of future work. 
\section{Experiments}
Our final {\ourdataset} comprises 10 different yearly snapshots and we evaluate entity linking (EL) performance on each of them individually. This evaluation setup allows us to 
study the effect of temporal corpus changes and assess the impact of increasing time lapses between the data used for model training and that on which the EL model is deployed
\citep{he2018time, agarwal2022temporal, luu2021time}. 
We 
train 
a bi-encoder baseline EL model (detailed in \secref{sec:baseline}) 
on the temporal snapshots from 2014 to 2022 separately and
then evaluate EL performance using the test sets of both past and future snapshots.

More specifically, our experiments
aim
to answer the following research questions:
\begin{enumerate*}[label=\textbf{(Q\arabic*)}]
    \item \label{it:q1-temp-degradation}
    Does a fixed entity linking (EL) model's performance degrade when applied to newer content?
    \item \label{it:q2-finetuning} 
    How does finetuning an EL model on more recent training data affect its performance on both old and newer content?
    \item \label{it:q3-emerging} 
    How does EL performance differ for resolving \emph{new} versus \emph{continual} entities?
\end{enumerate*}

\subsection{Baseline}
\label{sec:baseline}
\begin{figure}[t]
\begin{subfigure}{.32\textwidth}
  \centering
  \includegraphics[width=1.0\linewidth, trim={0.0cm 0.0cm 0.0cm 1.00cm},clip]{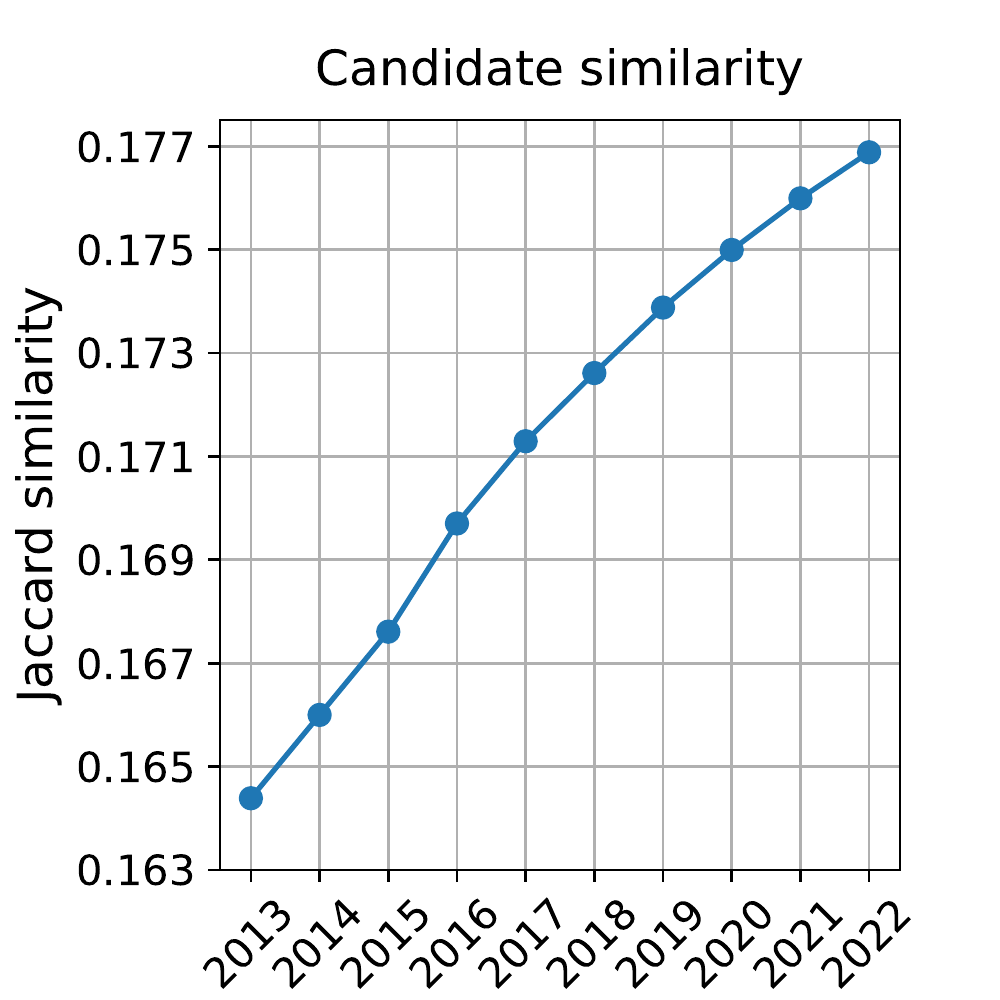}  
  \caption{Similarity between candidates returned by the bi-encoder baseline.}
  \label{fig:candidate_similarity}
\end{subfigure}\hfill
\begin{subfigure}{.32\textwidth}
  \centering
  \includegraphics[width=1.0\linewidth, trim={0.0cm 0.0cm 0.0cm 1.00cm},clip]{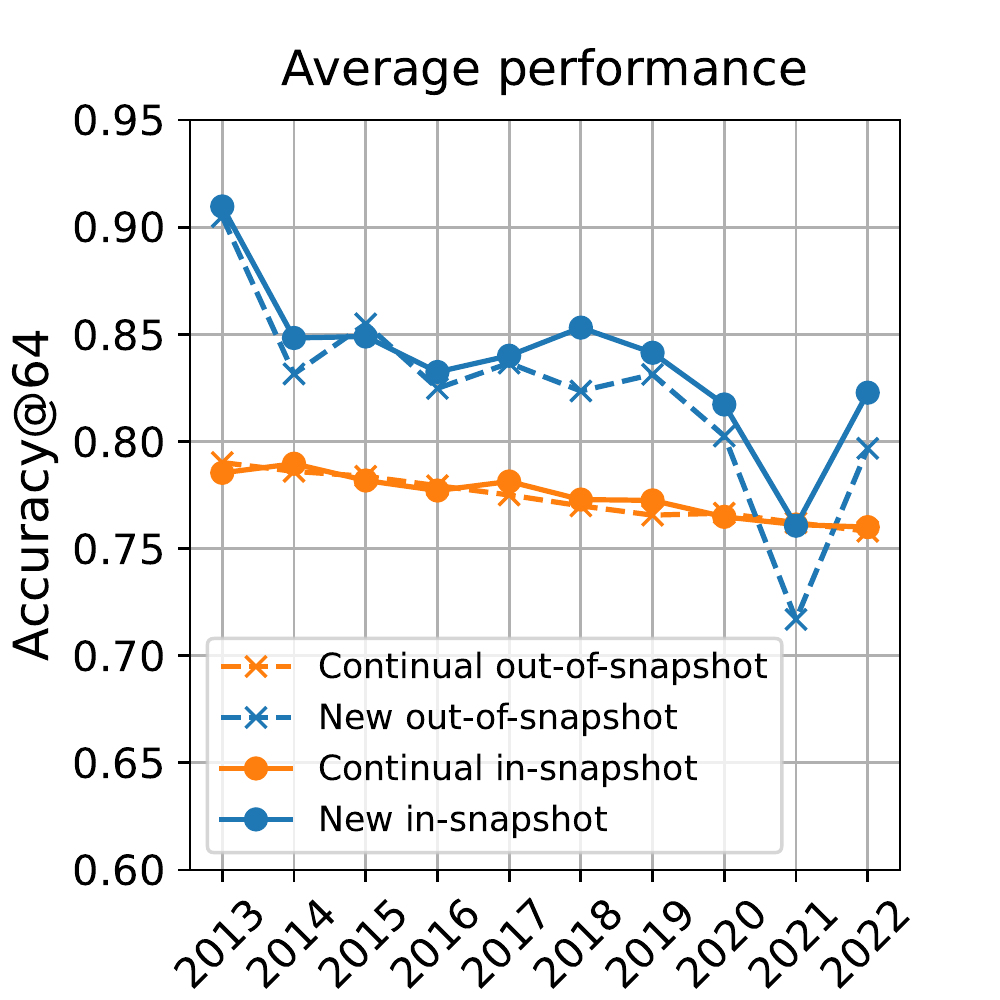}
  \caption{Difference in performance between \emph{new} and \emph{continual} entities.}
  \label{fig:avg_performance}
\end{subfigure}\hfill
\begin{subfigure}{.32\textwidth}
  \centering
  \includegraphics[width=1.0\linewidth, trim={0.0cm 0.0cm 0.0cm 1.00cm},clip]{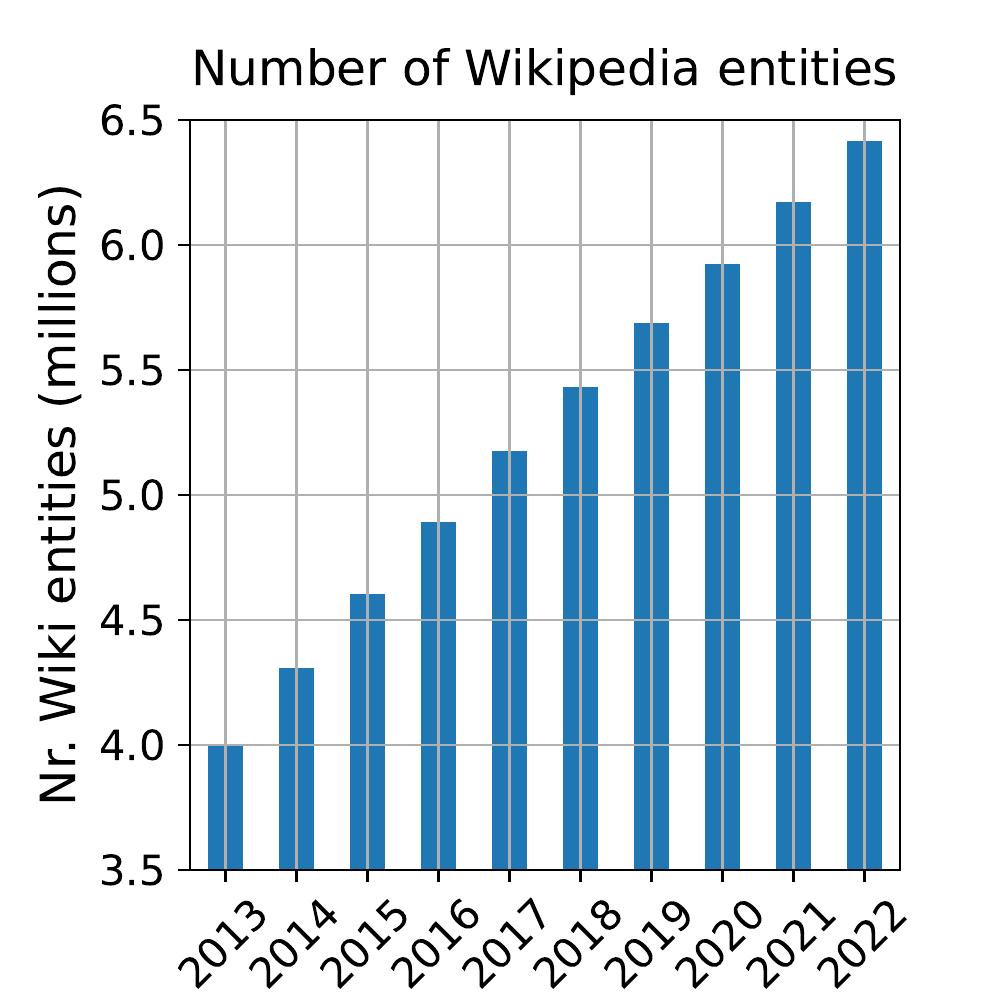}  
  \caption{Evolution of the number of entities in the Wikipedia KB.}
  \label{fig:nr_wiki_entities}
\end{subfigure}
\caption{Statistics related to the analysis of the results (\secref{sec:results_and_analysis}) across the temporal snapshots (x-axis).}
\label{fig:candidates_and_performance}
\end{figure}
We experiment with the bi-encoder \citep{mazare2018training,dinan2018wizard} baseline introduced in the BLINK model \cite{wu2019zero}. This method independently encodes the mention contexts 
from 
the entity descriptions, and then performs the retrieval in a dense space \citep{karpukhin2020dense} by matching the context of each mention with the closest candidate entities.  
For the entity description, we concatenate the title to the content of the page describing a particular entity. Both mention context as well as entity descriptions are truncated to 128 BERT tokens as per BLINK model \citep{wu2019zero}.
Similarly to \cite{agarwal2022temporal, loureiro2022timelms},
we start from a pre-trained BERT model,\footnote{We use BERT-large, which is trained on a Wikipedia snapshot from 2018 \citep{joshi2019bert}.} which we finetune using our {\ourdataset} snapshots' training data --- rather than fully re-training the BERT language model on the respective year's full Wikipedia corpus.
We leave the latter full-fledged BERT (re-)training approach for future work.

\subsection{Results and analysis}
\label{sec:results_and_analysis}
\begin{table}[t] 
\caption{Accuracy@64 for \emph{continual} (top) and \emph{new} (bottom) entities. The intensity of colors is set on a row-by-row basis and indicates whether performance is \textcolor{darkspringgreen!100}{\textbf{better}} or \textcolor{deepcarmine!100}{\textbf{worse}} compared to the year the model was finetuned on (\ie the values that form the white diagonal).} 
\label{tab:general_table_shared}
\vspace{.75\baselineskip}
\centering 
\resizebox{\columnwidth}{!} 
{\begin{tabular}{c cccccccccc} 
	 \toprule 
	 \multicolumn{11}{c}{Continual Entities} \\ 
	 \midrule 
	 \backslashbox{\textbf{Train}}{\textbf{Test}}& 2013 & 2014 & 2015 & 2016 & 2017 & 2018 & 2019 & 2020 & 2021 & 2022 \\ 
	\midrule 
2013  & 0.785 & \cellcolor{deepcarmine!16}0.782 & \cellcolor{deepcarmine!23}0.778 & \cellcolor{deepcarmine!34}0.772 & \cellcolor{deepcarmine!40}0.769 & \cellcolor{deepcarmine!52}0.762 & \cellcolor{deepcarmine!60}0.758 & \cellcolor{deepcarmine!60}0.758 & \cellcolor{deepcarmine!68}0.754 & \cellcolor{deepcarmine!75}0.750 \\ 
2014  & \cellcolor{darkspringgreen!16}0.792 & 0.790 & \cellcolor{deepcarmine!20}0.785 & \cellcolor{deepcarmine!28}0.781 & \cellcolor{deepcarmine!37}0.777 & \cellcolor{deepcarmine!49}0.771 & \cellcolor{deepcarmine!59}0.767 & \cellcolor{deepcarmine!58}0.767 & \cellcolor{deepcarmine!67}0.763 & \cellcolor{deepcarmine!75}0.760 \\ 
2015  & \cellcolor{darkspringgreen!20}0.786 & \cellcolor{darkspringgreen!14}0.784 & 0.782 & \cellcolor{deepcarmine!23}0.777 & \cellcolor{deepcarmine!32}0.773 & \cellcolor{deepcarmine!44}0.769 & \cellcolor{deepcarmine!53}0.765 & \cellcolor{deepcarmine!57}0.764 & \cellcolor{deepcarmine!67}0.760 & \cellcolor{deepcarmine!75}0.757 \\ 
2016  & \cellcolor{darkspringgreen!43}0.789 & \cellcolor{darkspringgreen!29}0.784 & \cellcolor{darkspringgreen!22}0.781 & 0.777 & \cellcolor{deepcarmine!21}0.773 & \cellcolor{deepcarmine!36}0.768 & \cellcolor{deepcarmine!49}0.763 & \cellcolor{deepcarmine!49}0.763 & \cellcolor{deepcarmine!64}0.758 & \cellcolor{deepcarmine!75}0.755 \\ 
2017  & \cellcolor{darkspringgreen!54}0.794 & \cellcolor{darkspringgreen!43}0.791 & \cellcolor{darkspringgreen!35}0.788 & \cellcolor{darkspringgreen!22}0.785 & 0.781 & \cellcolor{deepcarmine!32}0.775 & \cellcolor{deepcarmine!45}0.771 & \cellcolor{deepcarmine!43}0.772 & \cellcolor{deepcarmine!59}0.768 & \cellcolor{deepcarmine!75}0.763 \\ 
2018  & \cellcolor{darkspringgreen!75}0.791 & \cellcolor{darkspringgreen!63}0.788 & \cellcolor{darkspringgreen!54}0.786 & \cellcolor{darkspringgreen!42}0.782 & \cellcolor{darkspringgreen!29}0.778 & 0.773 & \cellcolor{deepcarmine!24}0.769 & \cellcolor{deepcarmine!23}0.769 & \cellcolor{deepcarmine!41}0.764 & \cellcolor{deepcarmine!54}0.760 \\ 
2019  & \cellcolor{darkspringgreen!75}0.795 & \cellcolor{darkspringgreen!65}0.792 & \cellcolor{darkspringgreen!56}0.789 & \cellcolor{darkspringgreen!43}0.784 & \cellcolor{darkspringgreen!34}0.781 & \cellcolor{darkspringgreen!20}0.776 & 0.772 & \cellcolor{darkspringgreen!11}0.773 & \cellcolor{deepcarmine!24}0.767 & \cellcolor{deepcarmine!32}0.765 \\ 
2020  & \cellcolor{darkspringgreen!75}0.787 & \cellcolor{darkspringgreen!63}0.783 & \cellcolor{darkspringgreen!61}0.782 & \cellcolor{darkspringgreen!46}0.777 & \cellcolor{darkspringgreen!37}0.774 & \cellcolor{darkspringgreen!20}0.768 & \cellcolor{deepcarmine!10}0.765 & 0.765 & \cellcolor{deepcarmine!21}0.761 & \cellcolor{deepcarmine!35}0.756 \\ 
2021  & \cellcolor{darkspringgreen!75}0.788 & \cellcolor{darkspringgreen!67}0.785 & \cellcolor{darkspringgreen!59}0.782 & \cellcolor{darkspringgreen!47}0.777 & \cellcolor{darkspringgreen!38}0.773 & \cellcolor{darkspringgreen!27}0.769 & \cellcolor{darkspringgreen!17}0.764 & \cellcolor{darkspringgreen!17}0.764 & 0.761 & \cellcolor{deepcarmine!20}0.757 \\ 
2022  & \cellcolor{darkspringgreen!75}0.790 & \cellcolor{darkspringgreen!67}0.787 & \cellcolor{darkspringgreen!60}0.783 & \cellcolor{darkspringgreen!51}0.779 & \cellcolor{darkspringgreen!44}0.776 & \cellcolor{darkspringgreen!34}0.771 & \cellcolor{darkspringgreen!26}0.768 & \cellcolor{darkspringgreen!27}0.768 & \cellcolor{darkspringgreen!19}0.764 & 0.760 \\ 
	\midrule 
	 \multicolumn{11}{c}{New Entities} \\ 
	 \midrule 
	 \backslashbox{\textbf{Train}}{\textbf{Test}}& 2013 & 2014 & 2015 & 2016 & 2017 & 2018 & 2019 & 2020 & 2021 & 2022 \\ 
	\midrule 
2013  & 0.910 & \cellcolor{deepcarmine!36}0.819 & \cellcolor{deepcarmine!26}0.853 & \cellcolor{deepcarmine!34}0.826 & \cellcolor{deepcarmine!30}0.841 & \cellcolor{deepcarmine!38}0.812 & \cellcolor{deepcarmine!36}0.819 & \cellcolor{deepcarmine!44}0.791 & \cellcolor{deepcarmine!75}0.688 & \cellcolor{deepcarmine!49}0.774 \\ 
2014  & \cellcolor{darkspringgreen!36}0.908 & 0.848 & \cellcolor{darkspringgreen!16}0.862 & \cellcolor{deepcarmine!19}0.827 & \cellcolor{deepcarmine!12}0.843 & \cellcolor{deepcarmine!17}0.832 & \cellcolor{deepcarmine!12}0.842 & \cellcolor{deepcarmine!25}0.814 & \cellcolor{deepcarmine!75}0.704 & \cellcolor{deepcarmine!35}0.791 \\ 
2015  & \cellcolor{darkspringgreen!32}0.898 & \cellcolor{deepcarmine!21}0.823 & 0.849 & \cellcolor{deepcarmine!22}0.822 & \cellcolor{deepcarmine!28}0.808 & \cellcolor{deepcarmine!26}0.813 & \cellcolor{deepcarmine!17}0.832 & \cellcolor{deepcarmine!37}0.788 & \cellcolor{deepcarmine!75}0.706 & \cellcolor{deepcarmine!41}0.781 \\ 
2016  & \cellcolor{darkspringgreen!46}0.897 & \cellcolor{deepcarmine!10}0.832 & \cellcolor{darkspringgreen!26}0.862 & 0.832 & \cellcolor{darkspringgreen!13}0.839 & \cellcolor{deepcarmine!15}0.823 & \cellcolor{deepcarmine!15}0.823 & \cellcolor{deepcarmine!27}0.802 & \cellcolor{deepcarmine!75}0.718 & \cellcolor{deepcarmine!33}0.791 \\ 
2017  & \cellcolor{darkspringgreen!43}0.906 & \cellcolor{deepcarmine!13}0.832 & \cellcolor{darkspringgreen!18}0.857 & \cellcolor{deepcarmine!22}0.817 & 0.840 & \cellcolor{deepcarmine!18}0.824 & \cellcolor{deepcarmine!12}0.835 & \cellcolor{deepcarmine!35}0.791 & \cellcolor{deepcarmine!75}0.714 & \cellcolor{deepcarmine!26}0.808 \\ 
2018  & \cellcolor{darkspringgreen!38}0.908 & \cellcolor{deepcarmine!19}0.835 & \cellcolor{darkspringgreen!12}0.858 & \cellcolor{deepcarmine!21}0.830 & \cellcolor{deepcarmine!13}0.846 & 0.853 & \cellcolor{deepcarmine!19}0.835 & \cellcolor{deepcarmine!34}0.806 & \cellcolor{deepcarmine!75}0.728 & \cellcolor{deepcarmine!35}0.803 \\ 
2019  & \cellcolor{darkspringgreen!51}0.910 & \cellcolor{darkspringgreen!10}0.842 & \cellcolor{darkspringgreen!17}0.853 & \cellcolor{deepcarmine!22}0.821 & \cellcolor{darkspringgreen!10}0.842 & \cellcolor{darkspringgreen!10}0.843 & 0.841 & \cellcolor{deepcarmine!29}0.810 & \cellcolor{deepcarmine!75}0.734 & \cellcolor{deepcarmine!35}0.799 \\ 
2020  & \cellcolor{darkspringgreen!72}0.903 & \cellcolor{darkspringgreen!18}0.828 & \cellcolor{darkspringgreen!29}0.844 & \cellcolor{darkspringgreen!23}0.835 & \cellcolor{darkspringgreen!28}0.843 & \cellcolor{darkspringgreen!11}0.819 & \cellcolor{darkspringgreen!21}0.833 & 0.817 & \cellcolor{deepcarmine!75}0.728 & \cellcolor{deepcarmine!14}0.811 \\ 
2021  & \cellcolor{darkspringgreen!75}0.910 & \cellcolor{darkspringgreen!37}0.825 & \cellcolor{darkspringgreen!49}0.852 & \cellcolor{darkspringgreen!37}0.825 & \cellcolor{darkspringgreen!43}0.837 & \cellcolor{darkspringgreen!34}0.817 & \cellcolor{darkspringgreen!40}0.830 & \cellcolor{darkspringgreen!33}0.814 & 0.761 & \cellcolor{darkspringgreen!32}0.812 \\ 
2022  & \cellcolor{darkspringgreen!68}0.905 & \cellcolor{darkspringgreen!26}0.846 & \cellcolor{darkspringgreen!31}0.852 & \cellcolor{deepcarmine!11}0.820 & \cellcolor{darkspringgreen!14}0.830 & \cellcolor{darkspringgreen!14}0.830 & \cellcolor{darkspringgreen!16}0.832 & \cellcolor{deepcarmine!20}0.808 & \cellcolor{deepcarmine!75}0.732 & 0.823 \\ 
 
        \bottomrule 
        \end{tabular}} 
\end{table}
The results for \emph{continual} and \emph{new} entities are shown in \tabref{tab:general_table_shared}.
The rows thereof represent the snapshots
whose train set we used to finetune the bi-encoder model, while the columns indicate the snapshots test data each of the finetuned models was tested on.
The used metric is 
accuracy@64, 
which amounts to the fraction of anchor mentions in the test set for which the top-64 candidate entity list from the EL model includes the correct target.
We observe a consistent temporal decrease in performance for \emph{continual} entities \ref{it:q1-temp-degradation}. This is also 
reflected
in \figref{fig:avg_performance}, which 
illustrates
the average temporal degradation across all the finetuned models. 
We hypothesize that this degradation over time is because, as time evolves, the relative ``semantic distance'' between the ever growing number of entities shrinks: entities become harder to distinguish from one another.
In order to demonstrate this, we calculate the \textit{Jaccard Similarity} between consecutive descriptions of the top 64 candidate entities returned by the bi-encoder. We observe a consistent increase in this similarity metric illustrated in \figref{fig:candidate_similarity}. This growth in more similar entities is accompanied with a general increase in the number of entities in the Wikipedia KB (see \figref{fig:nr_wiki_entities}).
Consequently, the model is given an ever-increasing number of candidate target entities, which can potentially impact its performance.

\begin{figure}[t]
\begin{subfigure}{.32\textwidth}
  \centering
  \includegraphics[width=1.0\linewidth]{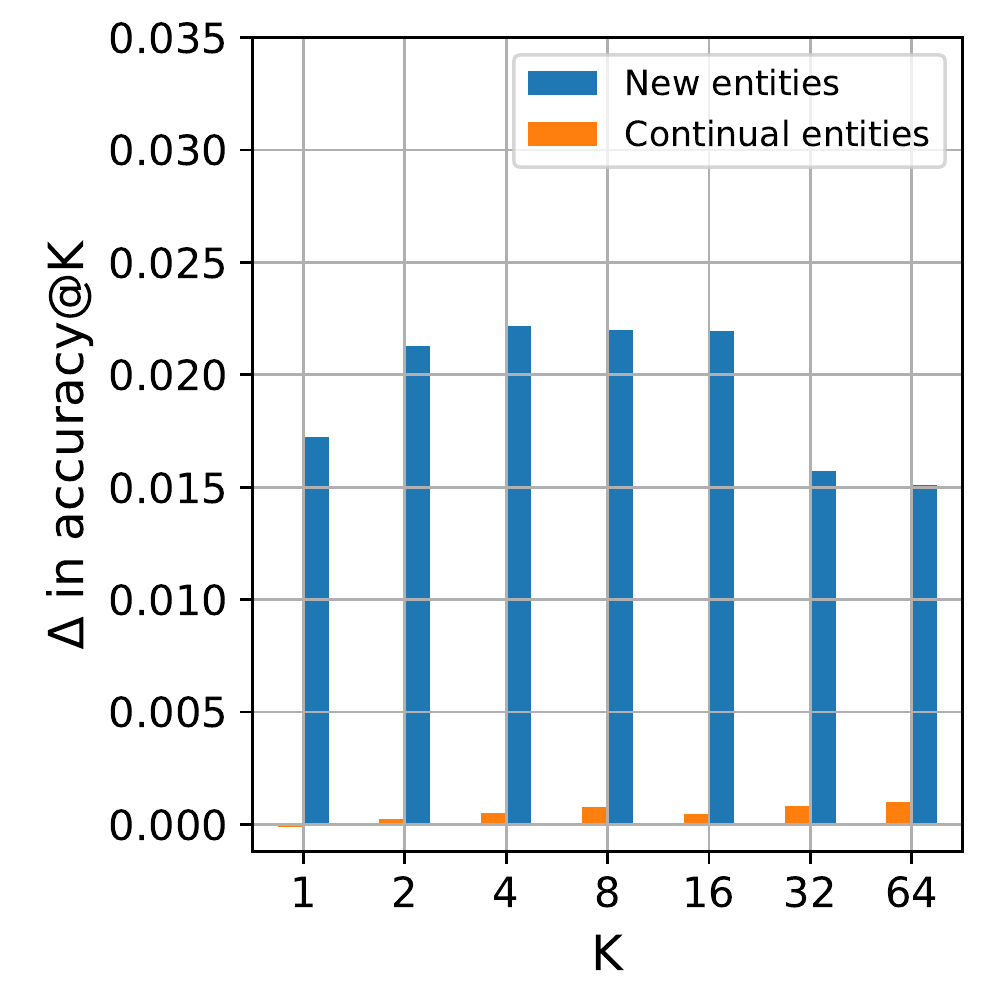}  
  \caption{\revklim{Effect of in-snapshot finetuning (y-axis) across different accuracy thresholds $K$.}}
  \label{fig:in-snapshot-k}
\end{subfigure}\hfill
\begin{subfigure}{.32\textwidth}
  \centering
  \includegraphics[width=1.0\linewidth]{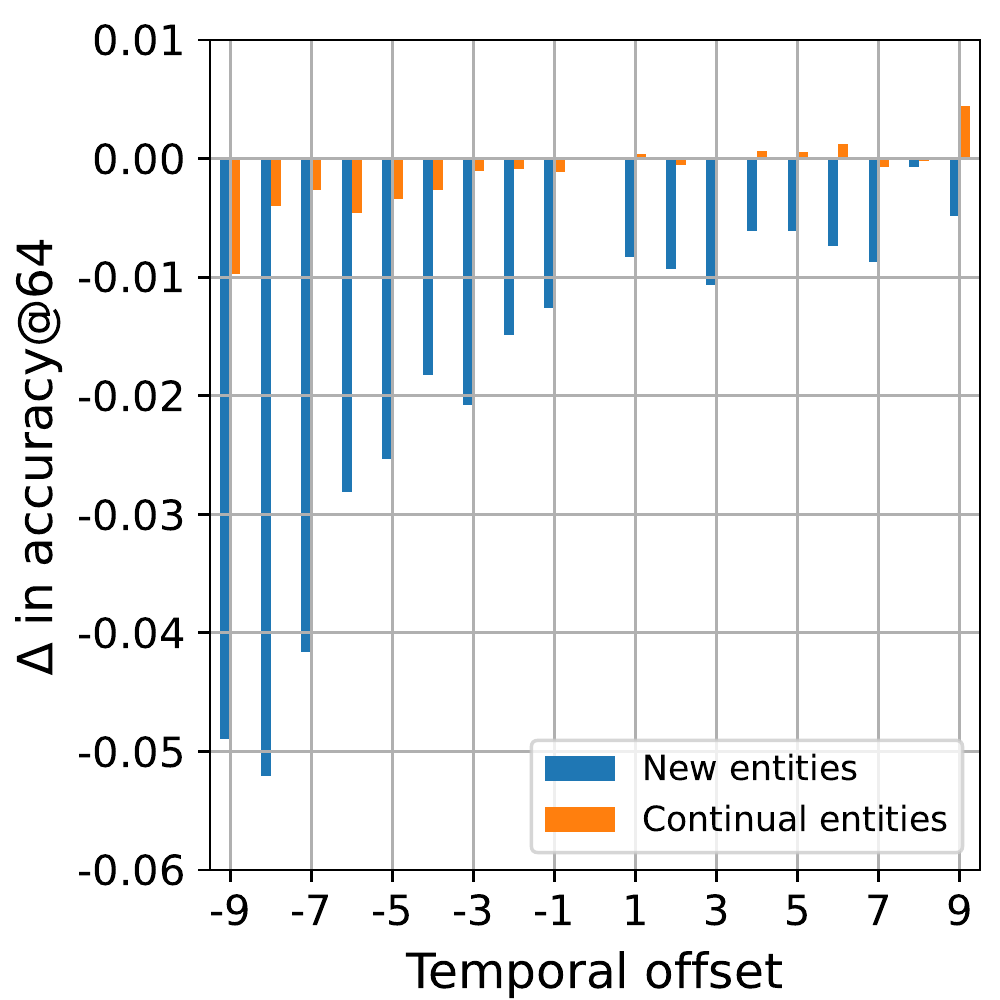}
  \caption{\revklim{In-snapshot finetuning (offset 0) compared to finetuning on past and future snapshots ($-$ and $+$ offsets).}}
  \label{fig:in-snapshot-offset}
\end{subfigure}\hfill
\begin{subfigure}{.32\textwidth}
  \centering
  \includegraphics[width=1.0\linewidth]{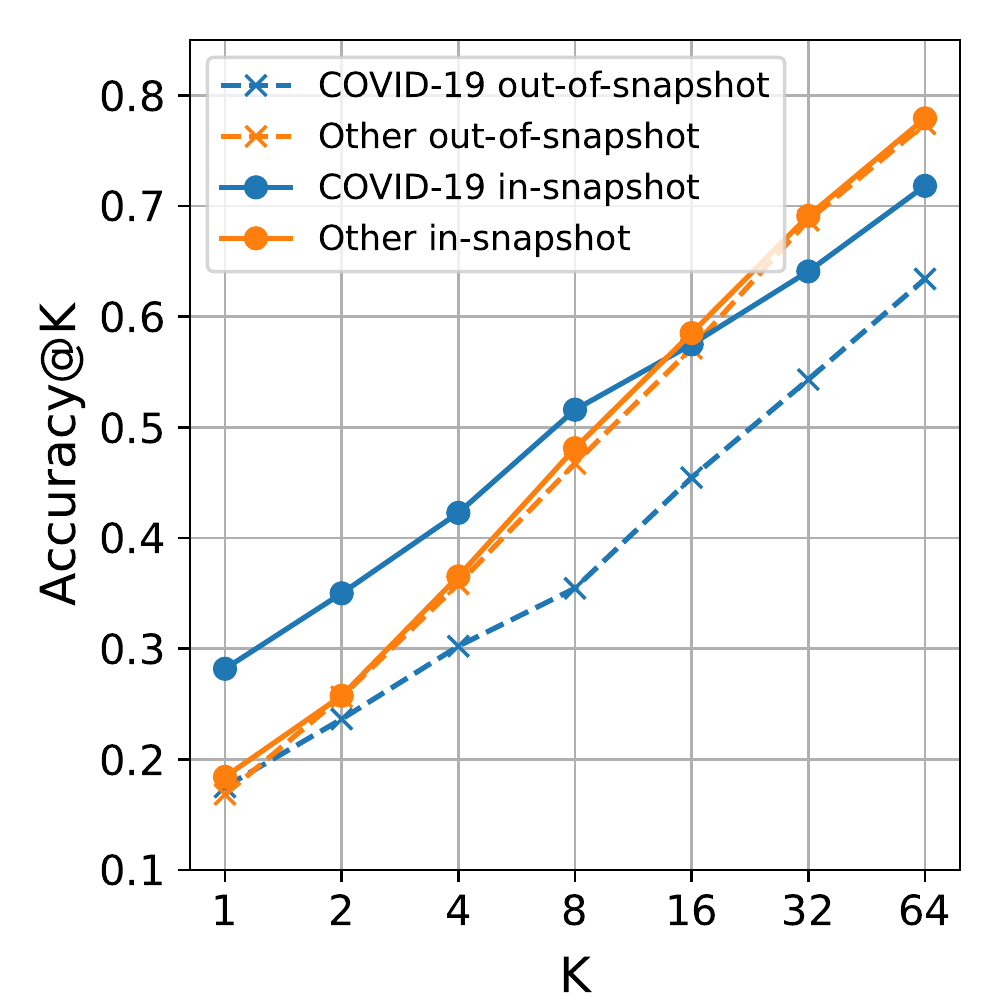}  
  \caption{\revklim{In-snapshot finetuning effect on COVID-19 related and other \textit{new entities} from 2021 snapshot.}}
  \label{fig:in-snapshot-covid}
\end{subfigure}
\caption[Impact of finetuning.]{\revklim{Impact of finetuning and evaluating on the same snapshot 
(\emph{in-snapshot}) 
compared to finetuning and evaluating on different snapshots
(\emph{out-of-snapshot}). 
We observe: 
\begin{enumerate*}[(a)]
    \item a superior impact of in-snapshot finetuning on \textit{new} entities compared to \textit{continual} ones,
    \item a decrease in performance when finetuning on increasingly older spanshots, 
    and 
    \item dominant effect of in-snapshot finetuning on entities that require fundamentally new knowledge (\eg COVID-19 related entities). 
\end{enumerate*}
}}
\label{fig:in_snapshot}
\end{figure}

\revklim{Furthermore, we analyze the impact finetuning on different snapshots has on the performance of the model \ref{it:q2-finetuning}. 
To this end, we distinguish between \textit{in-snapshot} and \textit{out-of-snapshot} 
finetuning setups. 
In \textit{in-snapshot} setup, 
the bi-encoder model is finetuned and evaluated
on the same snapshot. 
Conversely, in \textit{out-of-snapshot} setting, the model is evaluated on a different snapshot than the one used for its finetuning.
\Figref{fig:in-snapshot-k} illustrates the difference in performance between the in-snapshot and out-of-snapshot predictions for new and continual entities. 
We observe a general increase in performance for in-snapshot finetuning with a marginal gain for \textit{continual} entities compared to the \textit{new} ones.\footnote{\revklim{We analyze more in detail the difference in performance between \emph{new} and \textit{continual} entities in next paragraphs when addressing \ref{it:q3-emerging}.}} 
This general lower impact of in-snapshot finetuning on \textit{continual} entities, 
leads us to hypothesize that the actual knowledge needed to disambiguate 
most of these entities in {\ourdataset} changes very little with time. In order to 
verify this hypothesis, 
we randomly selected 100 
continual entity-mention pairs, 
and compared the difference in both mention contexts and entity descriptions between the years 2013 and 2022.
We found that in most cases (>95\%), while the textual description of the continual entity is changed (supported by \figsref{fig:evolution_entities_jaccard}{fig:evolution_entities_edistance}), its meaning remains the same.
}

\revklim{Moreover, we address the second part of \textbf{Q2}
targeting
the effect of timespan between the snapshot used for finetuning and the one used for evaluation.
To accomplish this, in \figref{fig:in-snapshot-offset} 
 we showcase 
the impact of in-snapshot finetuning relative to the \textit{temporal offset} between the snapshot the model was tested and the snapshot the model was finetuned on. For negative temporal offset,\footnote{\revklim{Evaluation snapshot comes from later time period than the snapshot the model was finetuned on.}} we observe a decrease in the performance difference 
between in-snapshot and out-of-snapshot setups
as the offset approaches to zero. This indicates that the model can benefit more from recent snapshots than from snapshots further in the past. 
Curiously, we 
observe a 
slight increase in performance for out-of-snapshot \textit{continual} entities trained on future snapshots (positive temporal offsets in \figref{fig:in-snapshot-offset}). 
This suggests that the changes in continual entities are \textit{accumulative} in Wikipedia,
with later versions of entity descriptions also including the information from the past. 
For instance,
we have observed that for entities describing people, the newly added information on the occupation (\eg soccer coach) is appended to the occupation description a person had in the past (\eg soccer player).}

Next, we analyze the EL performance on \emph{new} entities and whether they are differently affected than the \emph{continual} ones \ref{it:q3-emerging}.
\revklim{We plot the in-snapshot and out-of-snapshot average temporal change in accuracy@64 scores across all finetuned models for both types of entities in \figref{fig:avg_performance}}.
\revklim{We observe that, in general, the performance on \emph{new} entities is superior to that on \emph{continual} ones. Furthermore, as observed above, the performance gain from in-snapshot finetuning on new entities is superior compared to that on continual ones (supported by \figref{fig:avg_performance} and \figsref{fig:in-snapshot-k}{fig:in-snapshot-offset}). This difference suggests that new entities require a higher degree of additional snapshot-specific knowledge to be correctly disambiguated.
Additionally, the graph in \figref{fig:avg_performance} reveals that this delta in performance is 
larger
for more recent years (starting from 2018). We hypothesize 
that this behaviour is due to the fact that the used original BERT model\citep{devlin2019bert} has not been exposed to more recent new entities during pre-training. It also suggests a complementary effect between task-specific finetuning on \ourdataset~dataset and language model pre-training on larger corpora.}  

\revklim{Furthermore, 
to better understand the superior performance on new entities, 
we manually analyze 100 randomly selected
\emph{new} entities from our dataset.} 
\revklim{We found that a large 
majority ($\sim$90\%) of entities 
were either events that are 
recurrent in nature (\eg ``2018 BNP Paribas Open'') ($\sim$68\%) or extracts of already existing pages ($\sim$22\%).}
We conjecture\footnote{\revklim{See \secref{app:additional_results} of the supplementary material for further details on the performance on these different new entity types.}} 
that these entities require little additional knowledge
to be disambiguated, since either they already exist (as part of the content of other entities) or are very similar to already existing entities in Wikipedia. 
This contrasts sharply with the performance drop observed for \emph{new} entities in the temporal snapshot 2021, as exhibited in both \figref{fig:avg_performance} and \tabref{tab:general_table_shared}.
This decrease is mostly driven by COVID-19 related entities, which constitute 
\revklim{24\%}
of the new entities, which are linked to by 
\revklim{30\%}
of the mentions \revklim{in this spanshot}. 
The disambiguation of these cases requires completely new and 
fundamentally
different, previously non-existent knowledge.
\revklim{Since this knowledge is not present in the original corpus used to pre-train the BERT encoder nor in any of the previous snapshots,
our EL model based on it struggles.}

\revklim{Finally, we 
analyze the impact of new entities finetuning \ref{it:q2-finetuning} on the temporal snapshot 2021, 
for which our model exhibits the lowest temporal performance
 driven by COVID-19 disambiguation instances (see above)}. \revklim{\Figref{fig:in-snapshot-covid} showcases the 
impact of in- and out-of-snapshot finetuning
on the performance on
COVID-19 related entities compared to \textit{other} new entities 
for different thresholds $K$ of the accuracy@$K$ metric. 
We observe a large difference in performance 
(up to 14\% accuracy@64 points)
between COVID-19 related and the rest of the instances for out-of-snapshot finetuning.
This difference 
is significantly decreased when finetuning on the 2021 snapshot (in-snapshot finetuning), 
achieving
superior accuracy on COVID-19 related entities for lower values of $K$ compared to \textit{other} entities. In contrast, the difference between out- and in-snapshot performance on these non-COVID-19 related entities (\textit{other} entities in \figref{fig:in-snapshot-covid}) is marginal. 
This suggests that in-snapshot finetuning 
has dominant
impact on new entities that require fundamentally new, previously non-existent knowledge in Wikipedia. 
}

\section{Limitations and future work}
A number of dataset and model-related aspects were left unexplored in the current work.
Our clarifications thereof below may help the community to understand the limitations and 
potential future research directions to extend our efforts. 
\paragraph{Effect of pre-training on new corpora} Recent work has demonstrated the benefits of pre-training language models on more recent corpora (\eg the latest Wikipedia versions) when applied on downstream tasks \citep{agarwal2022temporal, loureiro2022timelms}. We hypothesize 
that this pre-training may also improve EL performance for our {\ourdataset},
especially for \emph{new} entities that require new 
world knowledge.
\paragraph{Changes in mention context} Our work focused mostly on changes in target entities, 
leaving the effect of changes in mention context on EL performance unexplored.
For example,
\figref{fig:evolution_mentions} 
shows a notable temporal drop in Jaccard vocabulary similarity of the context 
surrounding mentions.
This suggests that 
mentions, as well as the text surrounding them, are 
quite volatile and 
evolve over time, making them an interesting subject for future research. 

\paragraph{Cross-lingual time evolution} Our dataset is limited to English Wikipedia. Yet, since recent work \citep{botha2020entity, de2021multilingual} has shown the benefits of training EL models in a cross-lingual setting, 
studying cross-lingual temporal evolution of entity linking task 
may also
be an interesting future research direction. Furthermore, it will complement the recent growing interest in creating entity linking datasets for a number of low-resourced languages \citep{hennig2021mobie, ogrodniczuk2020wikipedia, caillaut2022automated,rosales2021towards}.

\section{Conclusion}
This paper introduced {\ourdataset}, a new large-scale 
temporal
entity linking 
dataset 
composed of 10 yearly snapshots of Wikipedia target entities linked to by anchor mentions. 
In our dataset creation pipeline, we put
special focus on the quality assurance and future extensibility of \ourdataset.
Furthermore, we 
established baseline entity linking results across different years,
which revealed a noticeable performance deterioration on test data more recent than the training data.
We further examined the most challenging cases, 
suggesting the need for updating the pre-trained language model of our EL model, at least to perform well on newly appearing entities that require new world knowledge (\eg in case of COVID-19).
Finally, we described limitations of our work and discussed potential future research directions.
\newpage
\begin{ack}
\revklim{\noindent Part of the research leading to these results has received funding from
\begin{enumerate*}[(i)]
\item the European Union's Horizon 2020 research and innovation programme under grant agreement no.\ 761488 for the CPN project,\footnote{\url{https://www.projectcpn.eu/}}
\item the Flemish Government under the programme ``Onderzoeksprogramma Artifici\"{e}le Intelligentie (AI) Vlaanderen'', 
 \item the Research Foundation -- Flanders grant no.\ V412922N for Long Stay Abroad at Copenhagen University, and
 \item DFF Sapere Aude grant No 0171-00034B ‘Learning to Explain Attitudes on Social Media (EXPANSE)’.
\end{enumerate*}}
\end{ack}
{
\small
\bibliographystyle{plain}
\bibliography{tempel_paper}
}

\clearpage
\section*{Checklist}

The checklist follows the references.  Please
read the checklist guidelines carefully for information on how to answer these
questions.  For each question, change the default \answerTODO{} to \answerYes{},
\answerNo{}, or \answerNA{}.  You are strongly encouraged to include a {\bf
justification to your answer}, either by referencing the appropriate section of
your paper or providing a brief inline description.  For example:
\begin{itemize}
  \item Did you include the license to the code and datasets? \answerYes{See the supplementary materials.}
\end{itemize}
Please do not modify the questions and only use the provided macros for your
answers.  Note that the Checklist section does not count towards the page
limit.  In your paper, please delete this instructions block and only keep the
Checklist section heading above along with the questions/answers below.

\begin{enumerate}

\item For all authors...
\begin{enumerate}
  \item Do the main claims made in the abstract and introduction accurately reflect the paper's contributions and scope?
    \answerYes{}
  \item Did you describe the limitations of your work?
    \answerYes{}
  \item Did you discuss any potential negative societal impacts of your work?
    \answerNA{}
  \item Have you read the ethics review guidelines and ensured that your paper conforms to them?
    \answerYes{}
\end{enumerate}

\item If you are including theoretical results...
\begin{enumerate}
  \item Did you state the full set of assumptions of all theoretical results?
    \answerNA{}
	\item Did you include complete proofs of all theoretical results?
    \answerNA{}
\end{enumerate}

\item If you ran experiments (e.g. for benchmarks)...
\begin{enumerate}
  \item Did you include the code, data, and instructions needed to reproduce the main experimental results (either in the supplemental material or as a URL)?
    \answerYes{The link to the dataset will be shared as part of the supplementary material.}
  \item Did you specify all the training details (e.g., data splits, hyperparameters, how they were chosen)?
    \answerYes{See the supplementary material.}  
	\item Did you report error bars (e.g., with respect to the random seed after running experiments multiple times)?
    \answerNo{No additional computational resources for this, yet the results across multiple temporal snapshots used to finetune are consistent. } 
	\item Did you include the total amount of compute and the type of resources used (e.g., type of GPUs, internal cluster, or cloud provider)?
    \answerYes{See the supplementary material.}  
\end{enumerate}

\item If you are using existing assets (e.g., code, data, models) or curating/releasing new assets...
\begin{enumerate}
  \item If your work uses existing assets, did you cite the creators?
    \answerYes{}
  \item Did you mention the license of the assets?
    \answerYes{See supplementary material} 
  \item Did you include any new assets either in the supplemental material or as a URL?
    \answerNo{}
  \item Did you discuss whether and how consent was obtained from people whose data you're using/curating?
    \answerNA{}
  \item Did you discuss whether the data you are using/curating contains personally identifiable information or offensive content?
    \answerNA{}
\end{enumerate}

\item If you used crowdsourcing or conducted research with human subjects...
\begin{enumerate}
  \item Did you include the full text of instructions given to participants and screenshots, if applicable?
    \answerNA{}
  \item Did you describe any potential participant risks, with links to Institutional Review Board (IRB) approvals, if applicable?
    \answerNA{}
  \item Did you include the estimated hourly wage paid to participants and the total amount spent on participant compensation?
    \answerNA{}
\end{enumerate}

\end{enumerate}
\clearpage
\appendix

\section{Supplementary material}
\subsection{Dataset and code distribution}
\paragraph{Link to the dataset} The reviewers can access the dataset using the following link: \url{https://cloud.ilabt.imec.be/index.php/s/RinXy8NgqdW58RW}. The dataset and the baseline code will be made publicly available in a dedicated GitHub repository upon acceptance.
\paragraph{License} \ourdataset~is distributed under Creative Commons Attribution-ShareAlike 4.0 International license (CC BY-SA 4.0).\footnote{\url{https://creativecommons.org/licenses/by-sa/4.0/}} 
\paragraph{Maintenance} The maintenance and extension to further temporal snapshots of \ourdataset~will be carried out by the authors of the paper. Additionally, we will make the code public to create potential new variations and extensions of \ourdataset~using a number of hyperparameters (see Sections \ref{app:hyperparameters} and \ref{app:dataset_extension} for further details). 

\subsection{Datasheet for \ourdataset}
In this section we provide a more detailed documentation of the dataset with the intended uses. We base ourselves on the datasheet proposed by \cite{gebru2021datasheets}. 
\subsubsection{Motivation}
\paragraph{For what purpose was the dataset created?}
The \ourdataset~dataset was created to evaluate how the temporal 
change
of anchor mentions and that of target Knowledge Base (KB; \ie modification or creation of new entities) affects the \textit{entity linking} (EL) task. This contrasts with the currently existing datasets \citep{usbeck2015gerbil,roder2018gerbil,sevgili2020neural,petroni2020kilt}, which are associated with a single version of the target KB such as the Wikipedia 2010 for the widely adopted CoNLL-AIDA\citep{hoffart2011robust} dataset. We expect that \ourdataset~will encourage research in devising new models and architectures that are robust to temporal changes both in mentions as well as in the target KBs. 

\paragraph{Who created the dataset and on behalf of which entity?}
The dataset is the result of joint effort involving researchers from the University of Copenhagen and Ghent University. 

\paragraph{Who funded the creation of the dataset?}
The creation of \ourdataset~was funded by the following grants: 
\begin{enumerate}
    \item FWO (Fonds voor Wetenschappelijk Onderzoek) long-stay abroad grant V412922N.
    \item The Flemish Government fund under the programme ``Onderzoeksprogramma Artifici\"{e}le Intelligentie (AI) Vlaanderen''.
\end{enumerate}

\subsubsection{Composition}
\paragraph{What do the instances that comprise the dataset represent?}
Each of the instances consists of a mention in Wikipedia linked to target entity, i.e., a Wikipedia page, with a set of attributes. 
The dataset is organized in 10 yearly temporal snapshots starting from 
January 1, 2013 until January 1, 2022. See \secref{app:mention_entity_attributes} for further details on the attributes associated with each of the instances of our \ourdataset~dataset. 

\paragraph{How many instances are there in total?}
\Tabref{tab:dataset_detail2} of the main manuscript summarizes the number of instances (\# Anchor Mentions) of each of the entity categories (\textit{continual} and \textit{new}) in \ourdataset. See \secref{app:mention_per_entity_distribution} for additional statistics on mention per entity distribution. 
\begin{figure}[!t]
\centering
\includegraphics[width=0.6\columnwidth]{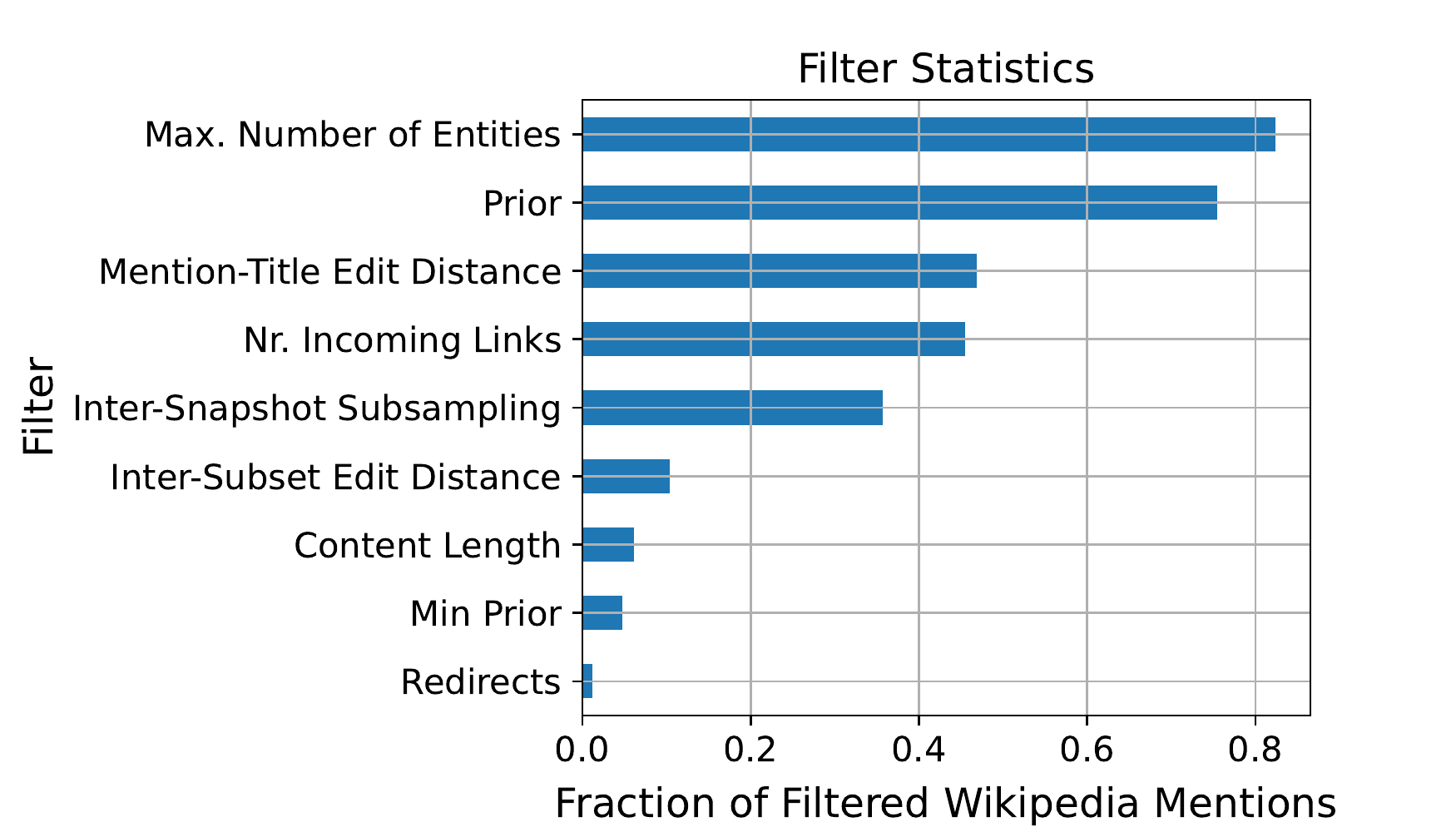}
\caption[todo]{Figure showcasing the fraction of filtered Wikipedia mentions by each of the filters executed during \ourdataset~generation. }
\label{fig:filter_effect}
\end{figure}
\paragraph{Does the dataset contain all possible instances or is it a sample
(not necessarily random) of instances from a larger set?}
\ourdataset~contains a sample of all the possible anchor mentions linked to target entities from Wikipedia. The following are the filters applied to obtain the instances in the final \ourdataset~dataset whose effect is also summarized in \figref{fig:filter_effect}:
\begin{enumerate}
    \item \textbf{Prior-based filtering}: we exclude all the mentions for which the correct entity it refers to has the highest \textit{prior}~\citep{yamada2016joint} as calculated in \equref{eq:mention_prior} of the manuscript. This filtering is done with the goal of creating a more challenging dataset. 

    \emph{Value to create \ourdataset}: mentions with mention prior rank $>$ 1 among other mentions referring to the same entity. 
    
    \emph{Percentage of filtered out instances}: between 74.20\% and 76.28\%, depending on the temporal snapshot. 
    
    \emph{Hyperparameter name:} \texttt{min\_men\_prior\_rank} (see \tabref{tab:hyperparameters} in \secref{app:hyperparameters}).
    
    \item \textbf{Entity relevance filtering}: we impose the restriction for target entity of having at least 10 incoming links (\ie at least 10 mentions linking to it) in order to be included in \ourdataset. Additionally, we filter out target entities whose description contains less than 10 tokens. This is done in order to avoid introducing potentially noisy and irrelevant entities that have not been sufficiently established by the Wikipedia community.  
    
    \emph{Value to create \ourdataset}: 10 for minimum number of incoming links and 10 for minimum content length (in number of tokens) of target entity.
    
    \emph{Percentage of filtered out instances}: 
    \begin{itemize}
    	\item Minimum number of incoming links: between 42.66\% and 48.32\%, depending on the temporal snapshot. 
    	\item Minimum content length: between 0.06\% and 0.95\% depending on the temporal snapshot. 
    \end{itemize}

    \emph{Hyperparameter names:} \texttt{min\_nr\_inlinks} for minimum number of incoming links and \texttt{min\_len\_target\_ent} for minimum number of content length tokens (see \tabref{tab:hyperparameters} in \secref{app:hyperparameters}).

    \item \textbf{Min prior subsampling}: the mentions with very low mention prior are filtered out from \ourdataset. This way, we avoid introducing too infrequent and potentially erroneous mentions to refer to a particular entity. 
    
    \emph{Value to create \ourdataset}: 0.0001
    
    \emph{Percentage of filtered out instances}: between 0.37\% and 0.61\%, depending on the snapshot. 
    
    \emph{Hyperparameter name:} \texttt{min\_men\_prior} (see \tabref{tab:hyperparameters} in \secref{app:hyperparameters}).
    \item \textbf{Minimum mentions per entity}: has similar effect as previously explained \textit{min prior subsampling} (see above) filter. We do not use it in the creation of \ourdataset, relying completely on the \textit{min prior subsampling} filter.
        
    \emph{Value to create \ourdataset}: 1
    
    \textit{Percentage of filtered out instances}: 0\%
    
    \emph{Hyperparameter name:} \texttt{min\_mens\_per\_ent} (see \tabref{tab:hyperparameters} in \secref{app:hyperparameters}).
    
    \item \textbf{Edit distance mention title}: filters out the anchor mentions that are very similar to target entity page. This way, we expect to reduce the trivial cases where the entity linking can be simply predicted by mapping the mention to the title of the target entity.
    
    \emph{Value to create \ourdataset}: 0.2 (normalized edit distance). 
    
    \textit{Percentage of filtered out instances}: between 44.85\% and 48.99\%, depending on the snapshot. 
    
    \emph{Hyperparameter name:} \texttt{ed\_men\_title} (see \tabref{tab:hyperparameters} in \secref{app:hyperparameters}).
    
    \item \textbf{Redirect filtering}: we filter out anchor mentions that point to redirect pages (pages without content redirecting to other pages in Wikipedia). 
    
    \textit{Percentage of filtered out instances}: between 1.02\% and 1.47\%, depending on the snapshot. 
    
    \item \textbf{Inter-subset filtering}: 
    we enforce normalized edit distance between the mentions in different subsets referring to the same target entity to be higher than 0.2. This entails that the entities in \ourdataset~are linked to at least by 3 mentions with different surface form. The main goal of this filter is to avoid mention-entity tuple memorization by the models \citep{onoe2020fine}. 
    
    \emph{Value to create \ourdataset}: 0.2 normalized edit distance between mentions in different subsets. 
    
    \textit{Percentage of filtered out instances}: 10\%.
    
    \emph{Hyperparameter name:} \texttt{ed\_men\_subsets} (see \tabref{tab:hyperparameters} in \secref{app:hyperparameters}).
    
    \item \textbf{Maximum number of entities}: we restrict the number of target entities to 10,000 for \emph{continual} instances. The reason behind this is to build a dataset of manageable size with a reasonable number of target entities to experiment with. 

    \emph{Value to create \ourdataset}: 10,000 for \emph{continual} entities.

    \emph{Percentage of filtered out instances}: 82\%. 

    \emph{Hyperparameter name:} \texttt{nr\_ct\_ents\_per\_cut} (see \tabref{tab:hyperparameters} in \secref{app:hyperparameters})
    
    \item \textbf{Maximum number of mentions per entity}: this filtering limits the number of mentions per entity in order for the dataset to not be dominated by most popular entities. Particularly, for test and evaluation subsets we limit the number of mentions per entity to 10. This way, we expect the accuracy scores to not be dominated by links to popular target entities (\ie entities with a big number of incoming links). The limit for training set is higher (500), since we want it to be representative of the real mention per entity distribution in Wikipedia. The effect of imposing this limits can be observed in \figref{fig:men_entity_distribution} for both \emph{continual} as well as \emph{new} entities represented by a significant leap in the mentions-per-entity curve, particularly noticeable for validation and test subsets. 

	\emph{Value to create \ourdataset}: 10 for validation and test subsets, 500 for the train subset.

    \emph{Percentage of filtered out instances}: for \emph{continual} instances, 84\% for validation and test subsets and 28\% for the train subset. For \emph{new} instances, 45\% for validation and test subsets and 0.3\% for the train subset. 

    \emph{Hyperparameter name:} \texttt{max\_mens\_per\_ent} (see \tabref{tab:hyperparameters} in \secref{app:hyperparameters}).
    
    \item \textbf{Inter-snapshot subsampling}: 
    finally, we enforce that the number of continual and new entities as well as the number of mentions stays the same across the temporal snapshots (see \tabref{tab:dataset_detail2}). We achieve this by performing a random mention subsampling in snapshots with higher number of mentions, weighted by the difference in the number of mentions-per-entity. This produces a very similar mention-entity distribution across the temporal snapshots (see \secref{app:mention_per_entity_distribution} for further details). 
    
    \emph{Percentage of filtered out instances}: between 5\% and 35\%, it increases for more recent temporal snapshots as they have more instances in Wikipedia.     
\end{enumerate}

We do not filter on any attribute that could potentially produce evident biases in \ourdataset~(\eg gender, geographic location of the entities, etc.).

\paragraph{What data does each instance consist of?}
Each instance of a snapshot consists of: 
\begin{enumerate}
    \item Cleaned contextual text surrounding the anchor mention from the Wikipedia snapshot. Furthermore, we include the bert-tokenized version of the text used in our baseline. 
    \item Cleaned textual description of the target entity taken from the Wikipedia snapshot. Furthermore, we include the bert-tokenized version of the text used in our baseline. 
    \item A set of additional attributes defining the anchor mention and target entity. 
\end{enumerate}
For more details about the attributes, see \secref{app:mention_entity_attributes}. Furthermore, concrete examples 
of \ourdataset's instances are showcased in \secref{app:examples}. 

\paragraph{Is there a label or target associated with each instance? }
Yes, the target entity is represented by the Wikipedia page id. Furthermore, we also pair it with Wikidata QID of the corresponding Wikidata entity. These targets correspond to the attributes \texttt{target\_page\_id} and \texttt{target\_qid} described in \tabref{tab:attributes} (see \secref{app:mention_entity_attributes} for further details). 

\paragraph{Is any information missing from individual instances? } No, all the instances should have a complete information corresponding to the content as well as to the attributes. 

\paragraph{Are relationships between individual instances made explicit? } Yes, the relations between each of the instances and the target entity are made explicit by means of \texttt{target\_page\_id} and \texttt{target\_qid} attributes (see \secref{app:mention_entity_attributes} for further details), which uniquely identify the id of the Wikipedia page describing a particular entity and the Wikidata entity respectively.

\paragraph{Are there recommended data splits (e.g., training, development/validation,
testing)? } Yes, the dataset is divided in train, validation and test subsets (see \tabref{tab:dataset_detail2} for the distribution). 

\paragraph{Are there any errors, sources of noise, or redundancies in the dataset?}
We have taken multiple measures to build a high quality dataset, minimizing the number of noise or other errors (see \secref{sec:quality} of the main manuscript). Yet, \ourdataset~is not 100\% error free, and contains a few errors mostly due to erroneous Wikitext edits by the Wikipedia users. 

\paragraph{Is the dataset self-contained, or does it link to or otherwise rely on external resources?}
Yes, the dataset is self contained and consists of: \begin{enumerate}
    \item Instances divided in train, validation and test subsets (see \tabref{tab:dataset_detail2}).
    \item A description of all the entities of each of the Wikipedia snapshots. These entities form the complete candidate pool used by the models to predict the correct target entity. \Figref{fig:nr_wiki_entities} of the main manuscript illustrates the temporal evolution in size of the number of candidate entities. 
\end{enumerate}

\paragraph{Does the dataset contain data that might be considered confidential?} No, Wikipedia is a public resource. 

\paragraph{Does the dataset contain data that, if viewed directly, might be offensive, insulting, threatening, or might otherwise cause anxiety?} No, we haven't detected instances of such characteristics in \ourdataset. 

\paragraph{Does the dataset identify any subpopulations (e.g., by age, gender)?}  
While there are articles on different subpopulations on Wikipedia, there is no emphasis of the dataset on identifying or annotating those. 

\paragraph{Is it possible to identify individuals (i.e., one or more natural persons), either directly or indirectly (i.e., in combination with other data) from the dataset? } Only based on their Wikipedia article, no editor information is retained.

\paragraph{Does the dataset contain data that might be considered sensitive in any way? } Wikipedia is overall a resource aiming to be factual, therefore we can exclude this concern for most instances of \ourdataset.

\subsubsection{Collection process}
\paragraph{How was the data associated with each instance acquired? } The textual data of the context of anchor mention and that of the description of the target entity is directly taken from the Wikipedia snapshots. Conversely, the attributes associated with each of the instances are calculated (see \secref{app:mention_entity_attributes} for further details).

\paragraph{What mechanisms or procedures were used to collect the data
(e.g., hardware apparatuses or sensors, manual human curation,
software programs, software APIs)? }
The dataset was collected using the Wikipedia dumps from February of 2022. We detail further on the aspects related to the preprocessing, cleaning and labeling of \ourdataset~instances in \secref{app:sheet_preprocessing_cleaning} of the datasheet.  

\paragraph{Who was involved in the data collection process (e.g., students, crowdworkers, contractors) and how were they compensated (e.g., how much were crowdworkers paid)?} The dataset was automatically generated based on existing Wikipedia articles. Therefore, no human intervention was needed for the dataset generation.

\paragraph{Over what timeframe was the data collected? }
The \ourdataset~dataset was collected from 10 yearly snapshots of Wikipedia starting from 
January 1, 2013 until January 1, 2022.

\paragraph{Were any ethical review processes conducted (e.g., by an institutional review board)?} N/A 

\subsubsection{Preprocessing/cleaning/labeling}
\label{app:sheet_preprocessing_cleaning}
\paragraph{Was any preprocessing/cleaning/labeling of the data done (e.g.,
discretization or bucketing, tokenization, part-of-speech tagging,
SIFT feature extraction, removal of instances, processing of missing values)?}
The Wikipedia history logs content is available exclusively in Wikitext markup format.\footnote{\url{https://en.wikipedia.org/wiki/Help:Wikitext}} In order to obtain cleaned text we proceed as follows: 
\begin{enumerate}
    \item We use MediaWiki API to process the templates which can not be parsed using regular expressions. For example, this is the case of the Wikitext template \texttt{Convert}, where the markup like ``\texttt{\{\{convert|37|mm|in|abbr=on\}\}}'' is converted to ``\texttt{1.5 in}''.  
    \item We use regular expressions to extract mentions and links. While this can also be done using online Wikitext parsing tools, we found that these did not account for all the corner cases of mention parsing such as the ones involving the \textit{pipe trick}.\footnote{\url{https://en.wikipedia.org/wiki/Help:Pipe_trick}}
    \item Finally, we use \texttt{mwparserfromhell}\footnote{\url{https://github.com/earwig/mwparserfromhell}} tool for parsing the rest of the Wikitext content. 
\end{enumerate}
Furthermore, our dataset files also contain BERT tokenization of the context around the mentions as well as the textual content of entities.

\paragraph{Was the ``raw'' data saved in addition to the preprocessed/cleaned/labeled
data (e.g., to support unanticipated future uses)? } 
Yes, the raw data containing the Wikipedia history logs was saved on our cloud server in the following link: \url{https://cloud.ilabt.imec.be/index.php/s/BF9SkmQG2Tdjw8o}.

\paragraph{Is the software that was used to preprocess/clean/label the data
available?} Yes, the software will be made public upon acceptance. 

\subsubsection{Uses}
\paragraph{Has the dataset been used for any tasks already? } Yes, in our submitted manuscript we describe a retriever bi-encoder baseline \citep{wu2019zero} (see \secref{sec:results_and_analysis}). 

\paragraph{Is there a repository that links to any or all papers or systems that use the dataset? } N/A

\paragraph{What (other) tasks could the dataset be used for?} The covered task is temporally evolving entity linking. 

\paragraph{Is there anything about the composition of the dataset or the way it was collected and preprocessed/cleaned/labeled that might impact future uses? } N/A

\paragraph{ Are there tasks for which the dataset should not be used? } N/A

\paragraph{Will the dataset be distributed to third parties outside of the entity (e.g., company, institution, organization) on behalf of which
the dataset was created?} Yes, the dataset is of public access. 

\paragraph{How will the dataset be distributed (e.g., tarball on website,
API, GitHub)? } The \ourdataset~dataset will be made public on a GitHub repository together with the code to generate it. The baseline code and models will also be made public on the same repository. Due to the size, the dataset files will be hosted on the cloud server that belongs to Internet Technology and Data Science Lab (IDLab) at Ghent University (\url{https://cloud.ilabt.imec.be/index.php/s/RinXy8NgqdW58RW}). 

\paragraph{When will the dataset be distributed?} The dataset will be publicly distributed upon the submission of the camera ready version of our manuscript. 

\paragraph{Will the dataset be distributed under a copyright or other intellectual property (IP) license, and/or under applicable terms of use (ToU)? } The \ourdataset~dataset will be distributed under Creative Commons Attribution-ShareAlike 4.0 International license (CC BY-SA 4.0). 

\paragraph{Have any third parties imposed IP-based or other restrictions on the data associated with the instances?} N/A

\paragraph{Do any export controls or other regulatory restrictions apply to the dataset or to individual instances? } N/A

\subsubsection{Maintenance}
\paragraph{Who will be supporting/hosting/maintaining the dataset?}
The maintenance and extension of \ourdataset~will be carried out by the authors of the paper. Additionally, we will make the code publicly available to create potential new variations of \ourdataset~using a number of hyperparameters (see
\secref{app:hyperparameters} and \secref{app:dataset_extension} for further details). 

The dataset files will be hosted on the cloud server that belongs to Internet Technology and Data Science Lab (IDLab) at Ghent University (\url{https://cloud.ilabt.imec.be/index.php/s/RinXy8NgqdW58RW}).

\paragraph{How can the owner/curator/manager of the dataset be contacted (e.g., email address)?}
The owners of the dataset can be contacted at the following e-mail address: \url{klim.zaporojets@ugent.be}. 

\paragraph{Is there an erratum?} No, there is no erratum yet. 

\paragraph{Will the dataset be updated (e.g., to correct labeling errors, add new instances, delete instances)?}
The \ourdataset~will be regularly updated with newer snapshots (see \secref{app:dataset_extension}). 
In circumstances such as labeling errors, we will release the fixed version of the dataset with the respective version number. The introduction of the new version will be communicated using the \ourdataset~GitHub repository. 

\paragraph{If the dataset relates to people, are there applicable limits on the
retention of the data associated with the instances (e.g., were the
individuals in question told that their data would be retained for
a fixed period of time and then deleted)?} N/A

\paragraph{Will older versions of the dataset continue to be supported/hosted/maintained?}
Yes, the older version of the dataset will continue to be supported and hosted. All the versions will be numbered and we will provide the link to access each of these versions on our cloud storage server. 

\paragraph{If others want to extend/augment/build on/contribute to the
dataset, is there a mechanism for them to do so?}
Yes, we provide the code and functionality to re-generate and extend the dataset with new temporal snapshots (see Sections \ref{app:hyperparameters} and \ref{app:dataset_extension}). Yet, it is the responsibility of the users to provide hosting and maintenance to the newly generated dataset variations. 

\subsection{Mentions per entity distribution}
\label{app:mention_per_entity_distribution}
\begin{figure}[!t]
\centering
\includegraphics[width=1.0\columnwidth, trim={0.0cm 0.0cm 0.0cm 0.0cm},clip]{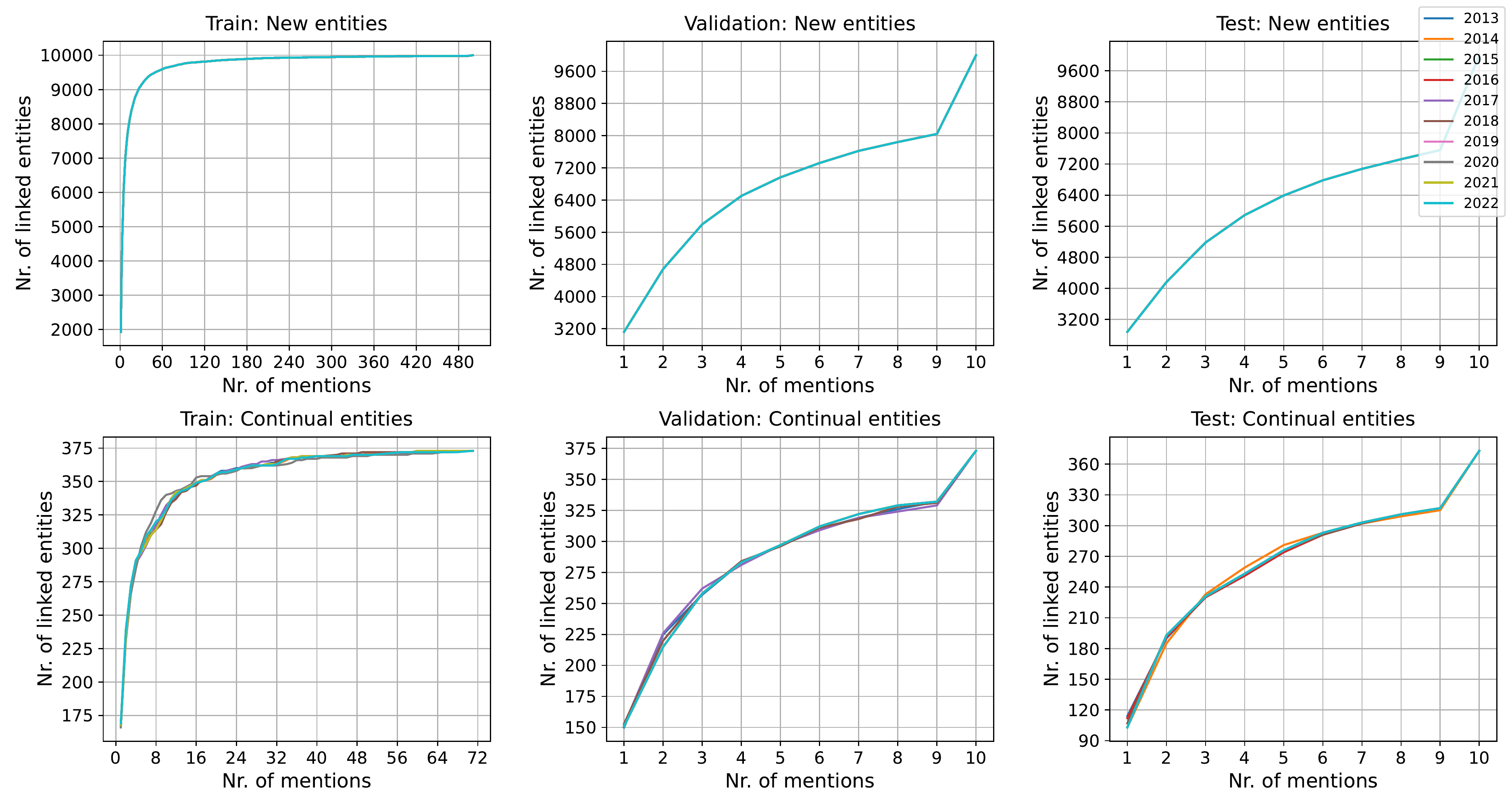}
\caption{Similar distribution of the data across the temporal snapshots (number of mentions per entity). 
This structurally unbiased setting enable to study exclusively the temporal effect
on the performance of the models for each of the different time periods. }
\label{fig:men_entity_distribution}
\end{figure}

\Figref{fig:men_entity_distribution} illustrates the similarity of mention per entity distribution across the temporal snapshots. This is achieved using weighted random subsampling so all the snapshots have equal number of instances (see \textit{Data Distributor} component description in \secref{sec:dataset_construction}). 
By enforcing this similarity between temporal snapshots, we ensure that the potential difference in the results is independent of cross-snapshot dataset distributional variations and only influenced by the dynamic temporal evolution of the content in \ourdataset. 
\subsection{Dataset creation hyperparameters}
\label{app:hyperparameters}
\begin{table}[t!]
\vspace{.75\baselineskip}
\centering
\caption{Hyperparameters that can be tuned during {\ourdataset} dataset creation. }
\vspace{.75\baselineskip}
\label{tab:hyperparameters}
{\renewcommand{\arraystretch}{1.2}
\begin{tabular}{p{0.24\linewidth}  p{0.55\linewidth} P{0.11\linewidth}}
        \toprule
         Hyperparamter & \multicolumn{1}{c}{Description} & \multicolumn{1}{c}{\ourdataset} \\ 
         \midrule 
         \texttt{snapshots} & Details (\eg timestamps) of the temporal snapshots to be generated. & 10 years \\
         \texttt{nr\_ct\_ents\_per\_cut} & Number of \textit{continual} entities per snapshot. & 10,000 \\
         \texttt{min\_mens\_per\_ent} & Minimum number of links a particular mention needs to have to target entity in order to be considered to be added in \ourdataset. & 1 \\
         \texttt{min\_men\_prior} & Minimum mention prior (see 
         \equref{eq:mention_prior} 
         in the main manuscript). & 0.0001 \\
         \texttt{max\_men\_prior} & Maximum mention prior. & 0.5 \\         
         \texttt{min\_men\_prior\_rank} & Minimum rank of mention prior among all the mentions pointing to a specific entity. & 2 \\
         \texttt{min\_ent\_prior} & Minimum entity prior as defined in \cite{yamada2016joint}: the ratio of links to the entity with respect to all of the links in the Wikipedia snapshot. & 0.0 \\
         \texttt{max\_ent\_prior} & Maximum entity prior. & 1.0 \\
         \texttt{min\_nr\_inlinks} & Minimum number of incoming links per entity. & 10 \\
         \texttt{min\_len\_target\_ent} & Minimum length of target entity page (in tokens). & 10 \\
         \texttt{max\_mens\_per\_ent} & Maximum number of mentions per entity. & 500/10/10\footnotemark \\
         \texttt{ed\_men\_title} & Minimum normalized edit distance between the mentions and the title of the target page they are linked to. & 0.2 \\
         \texttt{ed\_men\_subsets} & Minimum normalized edit distance between the mentions in different subsets linked to the same target entity.  & 0.2 \\
         \texttt{stable\_interval} & In seconds, the interval of time before the end of each snapshot from which the most stable version of Wikipedia has to be taken (see \secref{sec:quality} for further details).  & 2,592,000 (30 days) \\
         \revklim{\texttt{equal\_snapshots}} & \revklim{Whether the number of instances and the number of mentions per entity distribution is the same across the snapshots (see \secref{sec:quality} for further details). Equal cross-snapshot mention per entity distribution in \figref{fig:men_entity_distribution} is the result of setting this hyperparameter in True.}  & \revklim{True} \\         
         \bottomrule 
        \end{tabular}}
\end{table}
\footnotetext[5]{For train, validation and test sets respectively.}
\Tabref{tab:hyperparameters} summarizes the hyperparameters that can be tuned in order to automatically create the \ourdataset~dataset. This way, it is possible for the user to create different variation of the \ourdataset. The most relevant hyperparameter is \texttt{snapshots} that is used to specify the temporal intervals to create the snapshots. Below we detail two possible options we provide to specify such intervals. 
\paragraph{Option 1 - explicit snapshot specification} The user is expected to provide a list of timestamps in the format of \texttt{YYYY-MM-DDTHH:MM:SSZ}, each one defining a different snapshot. 

\paragraph{Option 2 - time span and interval} This option enables the user to define start and end dates of the time span from which the snapshots should be extracted. Furthermore, the interval value (\ie by using keywords such as ``weekly'' or specifying the interval in seconds)  has to also be specified. 

\subsection{Dataset extension}
\label{app:dataset_extension}
Additionally, we provide the option to extend the already existing dataset with new snapshots. Similarly as in the creation of new dataset (see \secref{app:hyperparameters} above), the \texttt{snapshots} hyperparameter is used to specify new snapshots which are then added 
to already existing \ourdataset~dataset. 

\subsection{Mention and entity attributes}
\label{app:mention_entity_attributes}
\clearpage
\begin{table}[t!]
\centering
\caption{Attributes associated to each of the mention-entity pairs for each of the temporal snapshots in \ourdataset. }
\vspace{.75\baselineskip}
\label{tab:attributes}
{\renewcommand{\arraystretch}{1.2}
\begin{tabular}{p{0.26\linewidth}  p{0.68\linewidth}}
        \toprule
         Attribute & \multicolumn{1}{c}{Description} \\ 
         \midrule 
         \texttt{subset} & The name of current subset (\ie train, validation or test).\\         
         \texttt{target\_page\_id} & The unique Wikipedia page id of the target entity. \\
        \texttt{target\_qid} & The unique Wikidata QID of the target entity. \\
         \texttt{snapshot} & The timestamp of the temporal snapshot from which the anchor mention and target entity attributes were extracted. \\
         \texttt{target} & The textual content of the target entity Wikipedia page. \\
         \texttt{target\_len} & The length in tokens of target Wikipedia page. \\         
         \texttt{target\_title} & The title of target entity Wikipedia page. \\
         \texttt{category} & Category of the target entity (\textit{new} or \textit{continual}). \\         
         \texttt{mention} & The text of the mention. \\
         \texttt{context\_left} & The textual context to the left of the mention. \\
         \texttt{context\_right} & The textual context to the right of the mention. \\
         \texttt{anchor\_len} & The length in tokens of the Wikipedia page where the anchor mention is located. \\
         \texttt{ed\_men\_title} & Normalized edit distance between the anchor mention and the title of the target Wikipedia page.  \\
         \texttt{overlap\_type} & Overlap type between the anchor mention and the target title as defined by \cite{logeswaran2019zero}. \\
         \texttt{men\_prior} & The mention prior (see \equref{eq:mention_prior} of the main manuscript). \\
         \texttt{men\_prior\_rank} & The rank of the current anchor mention compared to other mentions in Wikipedia pointing to target entity. \\
         \texttt{avg\_men\_prior} & The average value of prior of the mentions linked to the target entity in Wikipedia for \texttt{snapshot}. \\
         \texttt{ent\_prior} & Entity prior as defined in \cite{yamada2016joint}: the ratio of links to the entity with respect to all of the links in the Wikipedia snapshot. \\
         \texttt{nr\_inlinks} & Total number of incoming links to target entity. \\
         \texttt{nr\_dist\_mens} & Number of distinct (\ie with different surface form) mentions linked to target entity. \\
         \texttt{nr\_mens\_per\_ent} & Number of times the current mention appears in Wikipedia linked to target entity. \\ 
         \texttt{nr\_mens\_extracted} & Number of anchor mentions per current target entity in the \texttt{subset}. \\
         \texttt{anchor\_creation\_date} & The creation date (timestamp) of Wikipedia page where the anchor mention is located. \\
         \texttt{anchor\_revision\_date} & The timestamp of when the anchor Wikipedia page was last revised. \\
         \texttt{target\_creation\_date} & The timestamp of when the target Wikipedia entity page was created.  \\
         \texttt{target\_revision\_date} & The timestamp of when the target Wikipedia entity page was last modified. \\
         \bottomrule 
        \end{tabular}}
\end{table}
\clearpage
\Tabref{tab:attributes} describes the anchor mention and target entity related attributes present in \ourdataset. These attributes can be used to perform more in-depth analysis of the results.

\subsection{Baseline implementation details}
\label{app:training_hyperparameters}
We base our bi-encoder baseline model on the publicly available BLINK code.\footnote{\url{https://github.com/facebookresearch/BLINK}} We train all the models for 10 epochs with the learning rate of 1e-04 and the batch size of 64. We use AdamW optimizer with 10\% of warmup steps. Finally, we rely on \texttt{transformers} library \citep{wolf2020transformers} to get the pre-trained BERT-large representations.    All the experiments were run on NVIDIA V100 GPU with the following execution times: 
\begin{enumerate}
    \item \emph{Training}: 
    36 hours to train for 10 epochs per single snapshot.
    \item \emph{All Wikipedia entity encoding}: 7 days
    per finetuned model (on all the 10 Wikipedia snapshots) running on a single V100 GPU.  
    \item \emph{Evaluation}: 30 seconds per finetuned model per snapshot using FAISS \citep{johnson2019billion} library on GPU. 
\end{enumerate}

\subsection{Total amount of compute and the type of resources used to create \ourdataset}
In this section we provide the details on the computational resources used in each of the processing steps (see \secref{sec:dataset_construction} and  \figref{fig:el_pipeline} for further details) to create the \ourdataset~dataset: 

\begin{enumerate}
    \item \emph{Snapshot Data Extraction}: this processing step is responsible for creating the snapshots from the Wikipedia log files from February 1, 2022. This is a multi-processing step that is executed on a cluster with 80 CPUs and 110 GB of RAM and takes 5 days and 8 hours to complete. 
    \item \emph{Snapshot Dataset Building}: this is a multi-processing step that is executed on a cluster with 30 CPUs and 250 GB of RAM and takes 5 hours to complete. 
\end{enumerate}

\subsection{License of the assets}
We base the implementation of our baseline bi-encoder model on the publicly available BLINK \citep{wu2019zero} code. This asset is made available under MIT License (\url{https://opensource.org/licenses/MIT}).
\subsection{Examples}
\label{app:examples}
This section presents two illustrative examples of instances in \ourdataset. The first example contains the anchor mention linked to \textit{continual} entity, while the second one is the example of a link to \textit{new} entity. Both of the examples were taken from the snapshot of January 1, 2021. Furthermore, we trim the content length (\eg \texttt{target} attribute value) to only a few tokens for space reasons.

\subsubsection{Example 1: continual target entity}
\label{app:example1_continual}
\Tabref{tab:example1_continual} illustrates an example of the link to \textit{continual} target entity \textit{Sacramental\_bread}. It is worth noting that the creation date of this entity in Wikipedia (\texttt{target\_creation\_date} attribute) is of January 3, 2005. Yet, the version saved in the snapshot (\texttt{target\_revision\_date} attribute) is from December 30, 2020. \\ 
\clearpage
\begin{table}[t!]
\centering
\caption{Example of the instance corresponding to mention link to \textit{continual} entity (Sacramental\_bread created in 2005-01-03) in \ourdataset. }
\vspace{.75\baselineskip}
\label{tab:example1_continual}
{\renewcommand{\arraystretch}{1.2}
\begin{tabular}{p{0.26\linewidth}  p{0.68\linewidth}}
        \toprule
         Attribute & \multicolumn{1}{c}{Value} \\ 
         \midrule 
         \texttt{subset} & train \\
         \texttt{target\_page\_id} & 1359030 \\
         \texttt{target\_qid} & Q207104 \\
         \texttt{snapshot} & 2021-01-01T00:00:00Z \\
         \texttt{target} & ``Sacramental bread, sometimes called altar bread, Communion ...'' \\
         \texttt{target\_len} & 7,568 \\         
         \texttt{target\_title} & ``Sacramental\_bread''. \\
         \texttt{category} & continual \\         
         \texttt{mention} & ``host'' \\
         \texttt{context\_left} & ``... devotional image, portrait or other religious symbol (such as the'' \\
         \texttt{context\_right} & ``). Garland paintings were typically collaborations between a ...'' \\
         \texttt{anchor\_len} & 6,519 \\
         \texttt{ed\_men\_title} & 0.9411  \\
         \texttt{overlap\_type} & LOW\_OVERLAP \\
         \texttt{men\_prior} & 0.0750 \\
         \texttt{men\_prior\_rank} & 7 \\
         \texttt{avg\_men\_prior} & 0.6864 \\
         \texttt{ent\_prior} & 1.7790e-6 \\
         \texttt{nr\_inlinks} & 225 \\
         \texttt{nr\_dist\_mens} & 13 \\
         \texttt{nr\_mens\_per\_ent} & 79 \\ 
         \texttt{nr\_mens\_extracted} & 58 \\
         \texttt{anchor\_creation\_date} & 2009-09-25T21:09:07Z \\
         \texttt{anchor\_revision\_date} & 2020-10-04T16:15:13Z\\
         \texttt{target\_creation\_date} & 2005-01-03T17:41:14Z \\
         \texttt{target\_revision\_date} & 2020-12-30T12:38:50Z \\
         \bottomrule 
        \end{tabular}}
\end{table}
\clearpage
\subsubsection{Example 2: new target entity}
\label{app:example2_new_entity}
\Tabref{tab:example2_new} illustrates an example of the link to \textit{new} target entity \textit{COVID-19\_pandemic\_in\_Portland,\_Oregon}. It is worth noting that the creation date of this entity in Wikipedia (\texttt{target\_creation\_date} attribute) is of March 23, 2020, which belongs to the interval of the considered snapshot: from January 1, 2020 until January 1, 2021. \\ 
\begin{table}[t!]
\centering
\caption{Example of the instance corresponding to mention link to \textit{new} entity (COVID-19\_pandemic\_in\_Portland,\_Oregon created in 2020-03-23) in \ourdataset. }
\vspace{.75\baselineskip}
\label{tab:example2_new}
{\renewcommand{\arraystretch}{1.2}
\begin{tabular}{p{0.26\linewidth}  p{0.68\linewidth}}
        \toprule
         Attribute & \multicolumn{1}{c}{Value} \\ 
         \midrule 
         \texttt{subset} & train \\
         \texttt{target\_page\_id} & 63449958 \\
         \texttt{target\_qid} & Q88484856 \\
         \texttt{snapshot} & 2021-01-01T00:00:00Z \\
         \texttt{target} & ``The COVID-19 pandemic was confirmed to have reached ...'' \\
         \texttt{target\_len} & 26,432 \\         
         \texttt{target\_title} & ``COVID-19\_pandemic\_in\_Portland,\_Oregon'' \\
         \texttt{category} & new \\         
         \texttt{mention} & ``COVID-19 pandemic'' \\
         \texttt{context\_left} & ``Xico Xico and Xica both offered pickup service during the'' \\
         \texttt{context\_right} & ``, as of May 2020. '' \\
         \texttt{anchor\_len} & 2,437 \\
         \texttt{ed\_men\_title} & 0.5405  \\
         \texttt{overlap\_type} & AMBIGUOUS\_SUBSTRING \\
         \texttt{men\_prior} & 0.0009 \\
         \texttt{men\_prior\_rank} & 4 \\
         \texttt{avg\_men\_prior} & 0.2548 \\
         \texttt{ent\_prior} & 2.9255e-7 \\
         \texttt{nr\_inlinks} & 37 \\
         \texttt{nr\_dist\_mens} & 3 \\
         \texttt{nr\_mens\_per\_ent} & 23 \\ 
         \texttt{nr\_mens\_extracted} & 18 \\
         \texttt{anchor\_creation\_date} & 2020-12-08T00:23:50Z \\
         \texttt{anchor\_revision\_date} & 2020-12-09T15:41:18Z\\
         \texttt{target\_creation\_date} & 2020-03-23T04:22:55Z \\
         \texttt{target\_revision\_date} & 2020-11-16T03:59:06Z \\
         \bottomrule 
        \end{tabular}}
\end{table}

\subsection{Additional results}
\label{app:additional_results}
\begin{figure}[!t]
\centering
\includegraphics[width=0.6\columnwidth]{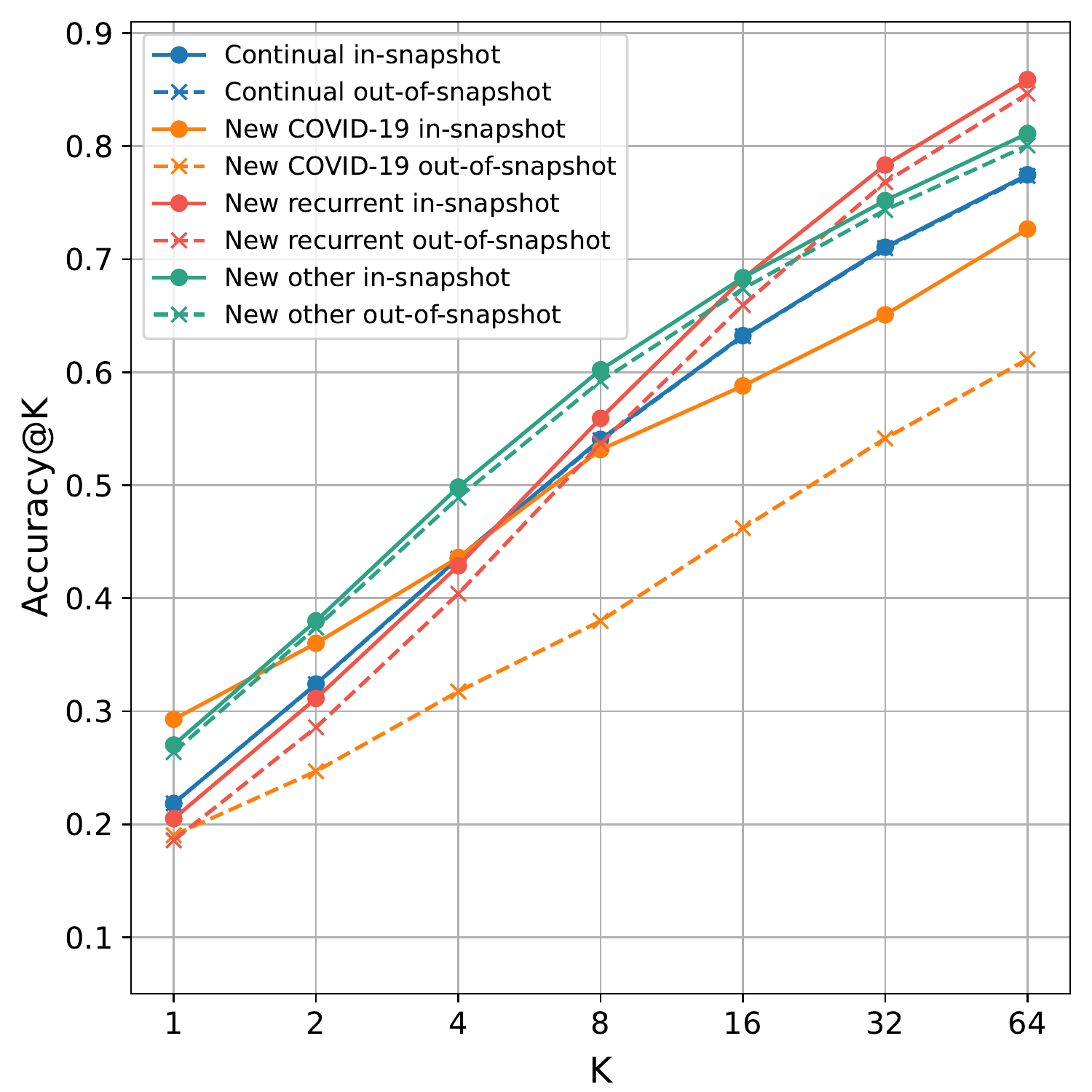}
\caption[Accuraccy@$K$ for different types of new entities]{\revklim{Accuraccy@$K$ for different values of $K \in \{1,2,4,8,16,32,64\}$. The results are grouped in four main categories: \begin{enumerate*}[(i)]
    \item mentions linked to \textit{continual} entities that exist in all of the \ourdataset~snapshots, 
    \item mentions linked to \textit{COVID-related} new entities (\ie with keywords such as ``COVID'' in target entity title), 
    \item mentions linked to \textit{recurrent} new entities (\ie entities representing events occurring periodically such as ``2018 BNP Paribas Open''), and 
    \item mentions linked to \textit{other} new entities. 
\end{enumerate*}}}
\label{fig:new_entity_types}
\end{figure}
Tables \ref{tab:general_table_shared_k_in_1}-\ref{tab:general_table_shared_k_in_64} present the results for different accuracy@$K$ for $K \in \{1,2,4,8,16,32,64\}$. 
\revklim{Furthermore, \figref{fig:new_entity_types} 
illustrates
the mean in- and out-of-snapshot (see \secref{sec:results_and_analysis} of the main manuscript) accuracy@$K$ performance across temporal snapshots on the following four target entity categories:}
\begin{enumerate}
    \item \revklim{\textit{Continual}: all the target \textit{continual} entities (\ie the entities that exist across all the temporal snapshots in \ourdataset~dataset).} 
    \item \revklim{\textit{COVID-19}: target \textit{new} entities that have COVID-related (\eg ``COVID'', ``coronavirus'', etc.) terms in the target entity title.}
    \item \revklim{\textit{Recurrent}: target \textit{new} entities whose titles contain the year and some of the keywords (\eg ``league'', ``election'', ``cup'', etc.) that indicate that an entity \revklim{is a repetitive event}
    (\eg ``2018 BNP Paribas Open'' which is part of \textit{yearly} BNB Paribas Open competitions).}
    \item \revklim{\textit{Other}: all the other target \textit{new} entities.}
\end{enumerate}
\revklim{The following are the main conclusions that can be drawn from the graph in \figref{fig:new_entity_types} that support or complement the findings described in \secref{sec:results_and_analysis} of the main manuscript:} 
\begin{enumerate}
    \item \revklim{New entities that require fundamentally new, previously non-existent knowledge to be disambiguated tend to have the lowest out-of-snapshot performance. This is the case of COVID-19 related disambiguation instances. These instances also experience the highest boost in performance when evaluated on in-snapshot setting (\ie the model is evaluated and finetuned on the same temporal snapshot).} 
    \item \revklim{The difference between in- and out-of-snapshot performances on \textit{continual} entities is the lowest. This is also supported by \figref{fig:avg_performance} and \figsref{fig:in-snapshot-k}{fig:in-snapshot-offset} in the main manuscript. This suggests that the actual knowledge needed to disambiguate most of the \textit{continual} entities in {\ourdataset} changes very little with time.}
    \item \revklim{The model has the highest accuracy@64 performance on \textit{recurrent} new entities. Yet, the performance on these entities drops sharply for lower values of $K$. We hypothesize that predicting the correct recurrent event gets more challenging as $K$ decreases because 
    of the large number of
    very similar candidates to pick from (\eg many ``BNP Paribas Open'' championships that only differ in very few details such as the date).}
    \item \revklim{The difference between in- and out-of-snapshot performance for \textit{other} new entities is lower than for \textit{recurrent} and \textit{COVID-19} related ones. This is driven by new entities that are derived from existing entities in Wikipedia (\ie their content is a copy of already 
    established
    entities). We hypothesize that the model requires little additional knowledge to disambiguate these entities. 
    Still,
    it is part of future work to study \textit{other} new entities more in detail in order to find cases that represent intrinsically new knowledge similar to the identified COVID-19 entity cluster.}
\end{enumerate}

\clearpage
\begin{table}[t] 
\caption{\textbf{Accuracy@1} for \emph{continual} (top) and \emph{new} (bottom) entities. The intensity of colors is set on a row-by-row basis and indicates whether performance is \textcolor{darkspringgreen!100}{\textbf{better}} or \textcolor{deepcarmine!100}{\textbf{worse}} compared to the year the model was finetuned on (\ie the values that form the white diagonal).} 
\vspace{.75\baselineskip}
\label{tab:general_table_shared_k_in_1} 
\centering 
\resizebox{\columnwidth}{!} 
{
\begin{tabular}{c cccccccccc} 
         \toprule 
         \multicolumn{11}{c}{Continual Entities} \\ 
         \midrule 
         \backslashbox{\textbf{Train}}{\textbf{Test}}& 2013 & 2014 & 2015 & 2016 & 2017 & 2018 & 2019 & 2020 & 2021 & 2022 \\ 
        \midrule 
2013  & 0.225 & \cellcolor{deepcarmine!21}0.219 & \cellcolor{deepcarmine!29}0.215 & \cellcolor{deepcarmine!26}0.217 & \cellcolor{deepcarmine!35}0.212 & \cellcolor{deepcarmine!46}0.206 & \cellcolor{deepcarmine!54}0.203 & \cellcolor{deepcarmine!54}0.203 & \cellcolor{deepcarmine!66}0.197 & \cellcolor{deepcarmine!75}0.192 \\ 
2014  & \cellcolor{darkspringgreen!17}0.229 & 0.226 & \cellcolor{deepcarmine!23}0.220 & \cellcolor{deepcarmine!19}0.221 & \cellcolor{deepcarmine!30}0.217 & \cellcolor{deepcarmine!40}0.212 & \cellcolor{deepcarmine!43}0.211 & \cellcolor{deepcarmine!52}0.207 & \cellcolor{deepcarmine!62}0.203 & \cellcolor{deepcarmine!75}0.197 \\ 
2015  & \cellcolor{darkspringgreen!37}0.228 & \cellcolor{darkspringgreen!22}0.223 & 0.219 & \cellcolor{darkspringgreen!10}0.219 & \cellcolor{deepcarmine!17}0.216 & \cellcolor{deepcarmine!32}0.211 & \cellcolor{deepcarmine!42}0.208 & \cellcolor{deepcarmine!46}0.206 & \cellcolor{deepcarmine!53}0.204 & \cellcolor{deepcarmine!75}0.196 \\ 
2016  & \cellcolor{darkspringgreen!37}0.230 & \cellcolor{darkspringgreen!26}0.227 & \cellcolor{darkspringgreen!13}0.222 & 0.221 & \cellcolor{deepcarmine!18}0.218 & \cellcolor{deepcarmine!30}0.214 & \cellcolor{deepcarmine!39}0.211 & \cellcolor{deepcarmine!48}0.208 & \cellcolor{deepcarmine!56}0.205 & \cellcolor{deepcarmine!75}0.199 \\ 
2017  & \cellcolor{darkspringgreen!58}0.240 & \cellcolor{darkspringgreen!46}0.237 & \cellcolor{darkspringgreen!21}0.229 & \cellcolor{darkspringgreen!20}0.229 & 0.226 & \cellcolor{deepcarmine!26}0.221 & \cellcolor{deepcarmine!34}0.219 & \cellcolor{deepcarmine!43}0.216 & \cellcolor{deepcarmine!60}0.211 & \cellcolor{deepcarmine!75}0.207 \\
\bottomrule 
        \end{tabular}} 
\end{table}
\begin{table}[t] 
\caption{\textbf{Accuracy@2} for \emph{continual} (top) and \emph{new} (bottom) entities. The intensity of colors is set on a row-by-row basis and indicates whether performance is \textcolor{darkspringgreen!100}{\textbf{better}} or \textcolor{deepcarmine!100}{\textbf{worse}} compared to the year the model was finetuned on (\ie the values that form the white diagonal).} 
\vspace{.75\baselineskip}
\label{tab:general_table_shared_k_in_2} 
\centering 
\resizebox{\columnwidth}{!} 
{\begin{tabular}{c cccccccccc} 
         \toprule 
         \multicolumn{11}{c}{Continual Entities} \\ 
         \midrule 
         \backslashbox{\textbf{Train}}{\textbf{Test}}& 2013 & 2014 & 2015 & 2016 & 2017 & 2018 & 2019 & 2020 & 2021 & 2022 \\ 
        \midrule 
2013  & 0.337 & \cellcolor{deepcarmine!20}0.330 & \cellcolor{deepcarmine!28}0.324 & \cellcolor{deepcarmine!31}0.322 & \cellcolor{deepcarmine!39}0.317 & \cellcolor{deepcarmine!49}0.311 & \cellcolor{deepcarmine!56}0.306 & \cellcolor{deepcarmine!61}0.302 & \cellcolor{deepcarmine!63}0.301 & \cellcolor{deepcarmine!75}0.293 \\ 
2014  & \cellcolor{darkspringgreen!17}0.339 & 0.335 & \cellcolor{deepcarmine!20}0.329 & \cellcolor{deepcarmine!21}0.328 & \cellcolor{deepcarmine!32}0.322 & \cellcolor{deepcarmine!41}0.317 & \cellcolor{deepcarmine!47}0.314 & \cellcolor{deepcarmine!54}0.310 & \cellcolor{deepcarmine!62}0.306 & \cellcolor{deepcarmine!75}0.299 \\ 
2015  & \cellcolor{darkspringgreen!36}0.339 & \cellcolor{darkspringgreen!23}0.333 & 0.327 & \cellcolor{deepcarmine!14}0.325 & \cellcolor{deepcarmine!20}0.323 & \cellcolor{deepcarmine!34}0.317 & \cellcolor{deepcarmine!46}0.312 & \cellcolor{deepcarmine!52}0.309 & \cellcolor{deepcarmine!60}0.305 & \cellcolor{deepcarmine!75}0.299 \\ 
2016  & \cellcolor{darkspringgreen!50}0.341 & \cellcolor{darkspringgreen!33}0.334 & \cellcolor{darkspringgreen!15}0.328 & 0.326 & \cellcolor{deepcarmine!19}0.322 & \cellcolor{deepcarmine!35}0.316 & \cellcolor{deepcarmine!40}0.314 & \cellcolor{deepcarmine!49}0.310 & \cellcolor{deepcarmine!60}0.306 & \cellcolor{deepcarmine!75}0.301 \\ 
2017  & \cellcolor{darkspringgreen!62}0.351 & \cellcolor{darkspringgreen!48}0.346 & \cellcolor{darkspringgreen!28}0.338 & \cellcolor{darkspringgreen!26}0.338 & 0.332 & \cellcolor{deepcarmine!21}0.328 & \cellcolor{deepcarmine!32}0.324 & \cellcolor{deepcarmine!43}0.320 & \cellcolor{deepcarmine!55}0.316 & \cellcolor{deepcarmine!75}0.309 \\
\bottomrule 
        \end{tabular}} 
\end{table}
\begin{table}[t] 
\caption{\textbf{Accuracy@4} for \emph{continual} (top) and \emph{new} (bottom) entities. The intensity of colors is set on a row-by-row basis and indicates whether performance is \textcolor{darkspringgreen!100}{\textbf{better}} or \textcolor{deepcarmine!100}{\textbf{worse}} compared to the year the model was finetuned on (\ie the values that form the white diagonal).} 
\vspace{.75\baselineskip}
\label{tab:general_table_shared_k_in_4} 
\centering 
\resizebox{\columnwidth}{!} 
{\begin{tabular}{c cccccccccc} 
         \toprule 
         \multicolumn{11}{c}{Continual Entities} \\ 
         \midrule 
         \backslashbox{\textbf{Train}}{\textbf{Test}}& 2013 & 2014 & 2015 & 2016 & 2017 & 2018 & 2019 & 2020 & 2021 & 2022 \\ 
        \midrule 
2013  & 0.449 & \cellcolor{deepcarmine!19}0.442 & \cellcolor{deepcarmine!24}0.439 & \cellcolor{deepcarmine!33}0.433 & \cellcolor{deepcarmine!41}0.428 & \cellcolor{deepcarmine!49}0.422 & \cellcolor{deepcarmine!57}0.417 & \cellcolor{deepcarmine!57}0.417 & \cellcolor{deepcarmine!66}0.411 & \cellcolor{deepcarmine!75}0.405 \\ 
2014  & \cellcolor{darkspringgreen!21}0.455 & 0.448 & \cellcolor{deepcarmine!19}0.443 & \cellcolor{deepcarmine!25}0.439 & \cellcolor{deepcarmine!36}0.433 & \cellcolor{deepcarmine!43}0.428 & \cellcolor{deepcarmine!51}0.424 & \cellcolor{deepcarmine!52}0.423 & \cellcolor{deepcarmine!63}0.416 & \cellcolor{deepcarmine!75}0.410 \\ 
2015  & \cellcolor{darkspringgreen!29}0.455 & \cellcolor{darkspringgreen!14}0.446 & 0.444 & \cellcolor{deepcarmine!21}0.438 & \cellcolor{deepcarmine!29}0.434 & \cellcolor{deepcarmine!41}0.427 & \cellcolor{deepcarmine!50}0.422 & \cellcolor{deepcarmine!50}0.422 & \cellcolor{deepcarmine!63}0.415 & \cellcolor{deepcarmine!75}0.409 \\ 
2016  & \cellcolor{darkspringgreen!46}0.453 & \cellcolor{darkspringgreen!30}0.446 & \cellcolor{darkspringgreen!21}0.442 & 0.437 & \cellcolor{deepcarmine!21}0.432 & \cellcolor{deepcarmine!35}0.426 & \cellcolor{deepcarmine!43}0.422 & \cellcolor{deepcarmine!44}0.422 & \cellcolor{deepcarmine!59}0.415 & \cellcolor{deepcarmine!75}0.408 \\ 
2017  & \cellcolor{darkspringgreen!75}0.464 & \cellcolor{darkspringgreen!58}0.458 & \cellcolor{darkspringgreen!46}0.454 & \cellcolor{darkspringgreen!26}0.448 & 0.443 & \cellcolor{deepcarmine!25}0.438 & \cellcolor{deepcarmine!38}0.434 & \cellcolor{deepcarmine!40}0.433 & \cellcolor{deepcarmine!55}0.428 & \cellcolor{deepcarmine!72}0.423 \\
\bottomrule 
        \end{tabular}} 
\end{table}
\begin{table}[t] 
\caption{\textbf{Accuracy@8} for \emph{continual} (top) and \emph{new} (bottom) entities. The intensity of colors is set on a row-by-row basis and indicates whether performance is \textcolor{darkspringgreen!100}{\textbf{better}} or \textcolor{deepcarmine!100}{\textbf{worse}} compared to the year the model was finetuned on (\ie the values that form the white diagonal).} 
\vspace{.75\baselineskip}
\label{tab:general_table_shared_k_in_8} 
\centering 
\resizebox{\columnwidth}{!} 
{\begin{tabular}{c cccccccccc} 
         \toprule 
         \multicolumn{11}{c}{Continual Entities} \\ 
         \midrule 
         \backslashbox{\textbf{Train}}{\textbf{Test}}& 2013 & 2014 & 2015 & 2016 & 2017 & 2018 & 2019 & 2020 & 2021 & 2022 \\ 
        \midrule 
2013  & 0.556 & \cellcolor{deepcarmine!17}0.551 & \cellcolor{deepcarmine!23}0.546 & \cellcolor{deepcarmine!32}0.539 & \cellcolor{deepcarmine!41}0.532 & \cellcolor{deepcarmine!49}0.526 & \cellcolor{deepcarmine!56}0.520 & \cellcolor{deepcarmine!57}0.520 & \cellcolor{deepcarmine!66}0.513 & \cellcolor{deepcarmine!75}0.507 \\ 
2014  & \cellcolor{darkspringgreen!16}0.563 & 0.559 & \cellcolor{deepcarmine!18}0.553 & \cellcolor{deepcarmine!29}0.546 & \cellcolor{deepcarmine!36}0.540 & \cellcolor{deepcarmine!48}0.532 & \cellcolor{deepcarmine!56}0.527 & \cellcolor{deepcarmine!58}0.526 & \cellcolor{deepcarmine!65}0.520 & \cellcolor{deepcarmine!75}0.514 \\ 
2015  & \cellcolor{darkspringgreen!26}0.561 & \cellcolor{darkspringgreen!16}0.555 & 0.552 & \cellcolor{deepcarmine!24}0.543 & \cellcolor{deepcarmine!29}0.540 & \cellcolor{deepcarmine!42}0.533 & \cellcolor{deepcarmine!54}0.526 & \cellcolor{deepcarmine!54}0.526 & \cellcolor{deepcarmine!63}0.520 & \cellcolor{deepcarmine!75}0.514 \\ 
2016  & \cellcolor{darkspringgreen!47}0.559 & \cellcolor{darkspringgreen!35}0.554 & \cellcolor{darkspringgreen!28}0.550 & 0.542 & \cellcolor{deepcarmine!20}0.537 & \cellcolor{deepcarmine!34}0.531 & \cellcolor{deepcarmine!48}0.524 & \cellcolor{deepcarmine!48}0.524 & \cellcolor{deepcarmine!61}0.518 & \cellcolor{deepcarmine!75}0.511 \\ 
2017  & \cellcolor{darkspringgreen!61}0.569 & \cellcolor{darkspringgreen!51}0.565 & \cellcolor{darkspringgreen!42}0.562 & \cellcolor{darkspringgreen!23}0.555 & 0.549 & \cellcolor{deepcarmine!30}0.542 & \cellcolor{deepcarmine!43}0.537 & \cellcolor{deepcarmine!43}0.537 & \cellcolor{deepcarmine!60}0.530 & \cellcolor{deepcarmine!75}0.525 \\
\bottomrule 
        \end{tabular}} 
\end{table}
\begin{table}[t] 
\caption{\textbf{Accuracy@16} for \emph{continual} (top) and \emph{new} (bottom) entities. The intensity of colors is set on a row-by-row basis and indicates whether performance is \textcolor{darkspringgreen!100}{\textbf{better}} or \textcolor{deepcarmine!100}{\textbf{worse}} compared to the year the model was finetuned on (\ie the values that form the white diagonal).} 
\vspace{.75\baselineskip}
\label{tab:general_table_shared_k_in_16} 
\centering 
\resizebox{\columnwidth}{!} 
{\begin{tabular}{c cccccccccc} 
         \toprule 
         \multicolumn{11}{c}{Continual Entities} \\ 
         \midrule 
         \backslashbox{\textbf{Train}}{\textbf{Test}}& 2013 & 2014 & 2015 & 2016 & 2017 & 2018 & 2019 & 2020 & 2021 & 2022 \\ 
        \midrule 
2013  & 0.648 & \cellcolor{deepcarmine!17}0.643 & \cellcolor{deepcarmine!21}0.639 & \cellcolor{deepcarmine!32}0.632 & \cellcolor{deepcarmine!40}0.626 & \cellcolor{deepcarmine!51}0.617 & \cellcolor{deepcarmine!58}0.613 & \cellcolor{deepcarmine!58}0.613 & \cellcolor{deepcarmine!68}0.605 & \cellcolor{deepcarmine!75}0.600 \\ 
2014  & \cellcolor{darkspringgreen!19}0.657 & 0.650 & \cellcolor{deepcarmine!15}0.647 & \cellcolor{deepcarmine!26}0.639 & \cellcolor{deepcarmine!32}0.635 & \cellcolor{deepcarmine!46}0.627 & \cellcolor{deepcarmine!53}0.622 & \cellcolor{deepcarmine!55}0.620 & \cellcolor{deepcarmine!67}0.613 & \cellcolor{deepcarmine!75}0.608 \\ 
2015  & \cellcolor{darkspringgreen!25}0.651 & \cellcolor{darkspringgreen!15}0.645 & 0.642 & \cellcolor{deepcarmine!21}0.636 & \cellcolor{deepcarmine!28}0.633 & \cellcolor{deepcarmine!45}0.624 & \cellcolor{deepcarmine!53}0.619 & \cellcolor{deepcarmine!54}0.619 & \cellcolor{deepcarmine!66}0.612 & \cellcolor{deepcarmine!75}0.608 \\ 
2016  & \cellcolor{darkspringgreen!41}0.652 & \cellcolor{darkspringgreen!30}0.646 & \cellcolor{darkspringgreen!23}0.643 & 0.637 & \cellcolor{deepcarmine!22}0.631 & \cellcolor{deepcarmine!41}0.621 & \cellcolor{deepcarmine!52}0.616 & \cellcolor{deepcarmine!54}0.615 & \cellcolor{deepcarmine!64}0.610 & \cellcolor{deepcarmine!75}0.605 \\ 
2017  & \cellcolor{darkspringgreen!71}0.660 & \cellcolor{darkspringgreen!54}0.655 & \cellcolor{darkspringgreen!46}0.652 & \cellcolor{darkspringgreen!28}0.646 & 0.640 & \cellcolor{deepcarmine!30}0.633 & \cellcolor{deepcarmine!45}0.628 & \cellcolor{deepcarmine!47}0.628 & \cellcolor{deepcarmine!66}0.621 & \cellcolor{deepcarmine!75}0.618 \\
\bottomrule 
        \end{tabular}} 
\end{table}
\begin{table}[t] 
\caption{\textbf{Accuracy@32} for \emph{continual} (top) and \emph{new} (bottom) entities. The intensity of colors is set on a row-by-row basis and indicates whether performance is \textcolor{darkspringgreen!100}{\textbf{better}} or \textcolor{deepcarmine!100}{\textbf{worse}} compared to the year the model was finetuned on (\ie the values that form the white diagonal).} 
\vspace{.75\baselineskip}
\label{tab:general_table_shared_k_in_32} 
\centering 
\resizebox{\columnwidth}{!} 
{\begin{tabular}{c cccccccccc} 
         \toprule 
         \multicolumn{11}{c}{Continual Entities} \\ 
         \midrule 
         \backslashbox{\textbf{Train}}{\textbf{Test}}& 2013 & 2014 & 2015 & 2016 & 2017 & 2018 & 2019 & 2020 & 2021 & 2022 \\ 
        \midrule 
2013  & 0.723 & \cellcolor{deepcarmine!16}0.719 & \cellcolor{deepcarmine!20}0.716 & \cellcolor{deepcarmine!31}0.710 & \cellcolor{deepcarmine!38}0.705 & \cellcolor{deepcarmine!50}0.697 & \cellcolor{deepcarmine!57}0.693 & \cellcolor{deepcarmine!59}0.692 & \cellcolor{deepcarmine!66}0.687 & \cellcolor{deepcarmine!75}0.682 \\ 
2014  & \cellcolor{darkspringgreen!18}0.731 & 0.727 & \cellcolor{deepcarmine!17}0.723 & \cellcolor{deepcarmine!26}0.717 & \cellcolor{deepcarmine!32}0.714 & \cellcolor{deepcarmine!46}0.706 & \cellcolor{deepcarmine!54}0.702 & \cellcolor{deepcarmine!55}0.702 & \cellcolor{deepcarmine!67}0.695 & \cellcolor{deepcarmine!75}0.690 \\ 
2015  & \cellcolor{darkspringgreen!22}0.727 & \cellcolor{darkspringgreen!14}0.723 & 0.721 & \cellcolor{deepcarmine!23}0.714 & \cellcolor{deepcarmine!31}0.710 & \cellcolor{deepcarmine!45}0.703 & \cellcolor{deepcarmine!52}0.700 & \cellcolor{deepcarmine!53}0.699 & \cellcolor{deepcarmine!65}0.693 & \cellcolor{deepcarmine!75}0.688 \\ 
2016  & \cellcolor{darkspringgreen!42}0.726 & \cellcolor{darkspringgreen!29}0.721 & \cellcolor{darkspringgreen!25}0.719 & 0.713 & \cellcolor{deepcarmine!20}0.709 & \cellcolor{deepcarmine!42}0.700 & \cellcolor{deepcarmine!52}0.696 & \cellcolor{deepcarmine!52}0.696 & \cellcolor{deepcarmine!63}0.692 & \cellcolor{deepcarmine!75}0.687 \\ 
2017  & \cellcolor{darkspringgreen!59}0.734 & \cellcolor{darkspringgreen!45}0.730 & \cellcolor{darkspringgreen!36}0.726 & \cellcolor{darkspringgreen!21}0.722 & 0.718 & \cellcolor{deepcarmine!32}0.710 & \cellcolor{deepcarmine!45}0.706 & \cellcolor{deepcarmine!46}0.706 & \cellcolor{deepcarmine!60}0.701 & \cellcolor{deepcarmine!75}0.696 \\
\bottomrule 
        \end{tabular}} 
\end{table}
\begin{table}[t] 
\caption{\textbf{Accuracy@64} for \emph{continual} (top) and \emph{new} (bottom) entities. The intensity of colors is set on a row-by-row basis and indicates whether performance is \textcolor{darkspringgreen!100}{\textbf{better}} or \textcolor{deepcarmine!100}{\textbf{worse}} compared to the year the model was finetuned on (\ie the values that form the white diagonal).} 
\vspace{.75\baselineskip}
\label{tab:general_table_shared_k_in_64} 
\centering 
\resizebox{\columnwidth}{!} 
{\begin{tabular}{c cccccccccc} 
         \toprule 
         \multicolumn{11}{c}{Continual Entities} \\ 
         \midrule 
         \backslashbox{\textbf{Train}}{\textbf{Test}}& 2013 & 2014 & 2015 & 2016 & 2017 & 2018 & 2019 & 2020 & 2021 & 2022 \\ 
        \midrule 
2013  & 0.785 & \cellcolor{deepcarmine!16}0.782 & \cellcolor{deepcarmine!23}0.778 & \cellcolor{deepcarmine!34}0.772 & \cellcolor{deepcarmine!40}0.769 & \cellcolor{deepcarmine!52}0.762 & \cellcolor{deepcarmine!60}0.758 & \cellcolor{deepcarmine!60}0.758 & \cellcolor{deepcarmine!68}0.754 & \cellcolor{deepcarmine!75}0.750 \\ 
2014  & \cellcolor{darkspringgreen!16}0.792 & 0.790 & \cellcolor{deepcarmine!20}0.785 & \cellcolor{deepcarmine!28}0.781 & \cellcolor{deepcarmine!37}0.777 & \cellcolor{deepcarmine!49}0.771 & \cellcolor{deepcarmine!59}0.767 & \cellcolor{deepcarmine!58}0.767 & \cellcolor{deepcarmine!67}0.763 & \cellcolor{deepcarmine!75}0.760 \\ 
2015  & \cellcolor{darkspringgreen!20}0.786 & \cellcolor{darkspringgreen!14}0.784 & 0.782 & \cellcolor{deepcarmine!23}0.777 & \cellcolor{deepcarmine!32}0.773 & \cellcolor{deepcarmine!44}0.769 & \cellcolor{deepcarmine!53}0.765 & \cellcolor{deepcarmine!57}0.764 & \cellcolor{deepcarmine!67}0.760 & \cellcolor{deepcarmine!75}0.757 \\ 
2016  & \cellcolor{darkspringgreen!43}0.789 & \cellcolor{darkspringgreen!29}0.784 & \cellcolor{darkspringgreen!22}0.781 & 0.777 & \cellcolor{deepcarmine!21}0.773 & \cellcolor{deepcarmine!36}0.768 & \cellcolor{deepcarmine!49}0.763 & \cellcolor{deepcarmine!49}0.763 & \cellcolor{deepcarmine!64}0.758 & \cellcolor{deepcarmine!75}0.755 \\ 
2017  & \cellcolor{darkspringgreen!54}0.794 & \cellcolor{darkspringgreen!43}0.791 & \cellcolor{darkspringgreen!35}0.788 & \cellcolor{darkspringgreen!22}0.785 & 0.781 & \cellcolor{deepcarmine!32}0.775 & \cellcolor{deepcarmine!45}0.771 & \cellcolor{deepcarmine!43}0.772 & \cellcolor{deepcarmine!59}0.768 & \cellcolor{deepcarmine!75}0.763 \\
\bottomrule 
        \end{tabular}} 
\end{table}

\end{document}